\documentclass[onecolumn]{article}

\usepackage{booktabs}
\usepackage{wrapfig}
\usepackage{multicol}
\usepackage[noadjust]{cite}

\usepackage{amsmath,amssymb,amsfonts}
\usepackage{graphicx}
\usepackage{textcomp}
\usepackage{times}
\usepackage{epsfig}
\usepackage{subcaption}
\usepackage{array,multirow}
\usepackage{float}
\usepackage{algorithm2e}
\usepackage{comment}
\usepackage[table,xcdraw]{xcolor}
\usepackage{hyperref}
\usepackage{soul}
\usepackage{authblk}
\usepackage[margin=0.7in]{geometry}
\usepackage{tabularx}
\usepackage{caption} 
\def\BibTeX{{\rm B\kern-.05em{\sc i\kern-.025em b}\kern-.08em
    T\kern-.1667em\lower.7ex\hbox{E}\kern-.125emX}}

\graphicspath{{./figures/}}
\DeclareGraphicsExtensions{.png , .jpg, .jpeg, .eps}

\newcommand{\ie}{\emph{i.e.}}
\newcommand{\eg}{\emph{e.g.}}
\newcommand{\etc}{\emph{etc}}
\newcommand{\etal}{\emph{et al.}}
\newcommand{\vs}{\emph{vs}}
\newcommand{\Fig}[1]{Fig. \ref{fig:#1}}
\newcommand{\Figure}[1]{Figure \ref{fig:#1}}

\newcommand{\ssSect}[1]{Sect. \ref{sssec:#1}}
\newcommand{\Tab}[1]{Table \ref{tab:#1}}
\newcommand{\Alg}[1]{Alg. \ref{alg:#1}}
\newcommand{\citetal}[2]{#1 {\etal} \cite{#2}}

\newcommand{\lab}{l}
\newcommand{\ulab}{u}
\newcommand{\datasets}{\mathcal{X}}

\newcommand{\image}{\mbox{I}}
\newcommand{\plab}{\hat{l}}
\newcommand{\XL}{\datasets^{\lab}}

\newcommand{\XU}{\datasets^{\ulab}}

\newcommand{\XP}{\datasets^{\plab}}

\newcommand{\CNN}{\Phi}
\newcommand{\OD}{\phi}
\newcommand{\HP}{\mathcal{H}}
\newcommand{\HPStop}{\HP_{{\tiny stp}}}
\newcommand{\HPSeq}{\HP_{{\tiny seq}}}
\newcommand{\HPOD}{\HP_{\CNN}}
\newcommand{\HPCoT}{\HP_{ct}}

\newcommand{\STOP}[4]{\mbox{{\small Stop?}}({#1},{#2},{#3},{#4})}
\newcommand{\Select}[3]{\mbox{{\small Slct}}({#1},{#2},{#3})}
\newcommand{\TD}[4]{\mbox{{\small Train}}({#1},{#2},{#3},{#4})}
\newcommand{\RD}[3]{\mbox{{\small Run}}({#1},{#2},{#3})}
\newcommand{\Fuse}[2]{\mbox{{\small Fuse}}({#1},{#2})}
\newcommand{\Rand}[3]{\mbox{{\small Rnd}}({#1},{#2}{#3})}

\newcommand{\smalldatasets}{\mathcal{S}}
\newcommand{\xL}{\smalldatasets^{\lab}}
\newcommand{\xP}{\smalldatasets^{\plab}}

\newcommand{\model}{\phi}
\newcommand{\domain}{\mathcal{D}}
\newcommand{\domainset}[1]{{\domain_{#1}}}
\newcommand{\train}{{tr}}
\newcommand{\test}{{tt}}

\newcommand{\Xtrain}{\datasets^{\train}}

\newcommand{\Xtest}{\datasets^{\test}}

\newcommand{\Rds}{\mathcal{R}}
\newcommand{\Vds}{\mathcal{V}}
\newcommand{\Rdspar}[1]{\Rds_{{\scriptsize \mbox{#1}}}}
\newcommand{\Vdspar}[1]{\Vds_{{\scriptsize \mbox{#1}}}}
\newcommand{\Kds}{\mathcal{K}}
\newcommand{\Wds}{\mathcal{W}}
\newcommand{\Kdstrain}{\Kds^{\train}}
\newcommand{\Wdstrain}{\Wds^{\train}}
\newcommand{\Kdstest}{\Kds^{\test}}
\newcommand{\Wdstest}{\Wds^{\test}}
\newcommand{\GAN}{\mathcal{G}}
\newcommand{\VGds}{\Vds_{\GAN}}
\newcommand{\GANKds}{{\GAN}_{\Kds}}
\newcommand{\VGKds}{\Vds_{\GANKds}}
\newcommand{\GANWds}{{\GAN}_{\Wds}}
\newcommand{\VGWds}{\Vds_{\GANWds}}
\newcommand{\GANRds}{{\GAN}_{\Rds}}

\newcommand{\VGRdspar}[1]{\Vds_{\GANRds,{\scriptsize \mbox{#1}}}}

\iftrue
\newcommand{\alertJW}[1]{{\color{magenta}{\bf JW:} #1}}
\else
\newcommand{\alertJW}[1]{}
\fi

\iftrue
\newcommand{\alertGV}[1]{{\color{blue}{\bf GV:} #1}}
\else
\newcommand{\alertGV}[1]{}
\fi

\providecommand{\keywords}[1]{\textbf{\textit{Index terms---}} #1}

\newcommand\blfootnote[1]{%
  \begingroup
  \renewcommand\thefootnote{}\footnote{#1}%
  \addtocounter{footnote}{-1}%
  \endgroup
}

\begin{document}

\title{Co-training for Deep Object Detection: Comparing Single-modal and Multi-modal Approaches}

\author[1,2]{\uppercase{Jose L. G\'{o}mez}}
\author[1]{\uppercase{Gabriel Villalonga}}
\author[1,2]{\uppercase{Antonio M. L\'opez}}

\affil[1]{Computer Vision Center (CVC) and Computer Science Dpt., Universitat Aut\`onoma de Barcelona (UAB), Spain.}
\affil[2]{Computer Science Department, Universitat Autònoma de Barcelona (UAB), 08193 Bellaterra, Spain.}


\maketitle

\begin{abstract}
Top-performing computer vision models are powered by convolutional neural networks (CNNs). Training an accurate CNN highly depends on both the raw sensor data and their associated ground truth (GT). Collecting such GT is usually done through human labeling, which is time-consuming and does not scale as we wish. This data labeling bottleneck may be intensified due to domain shifts among image sensors, which could force per-sensor data labeling. In this paper, we focus on the use of co-training, a semi-supervised learning (SSL) method, for obtaining self-labeled object bounding boxes (BBs), i.e., the GT to train deep object detectors. In particular, we assess the goodness of multi-modal co-training by relying on two different views of an image, namely, appearance (RGB) and estimated depth (D). Moreover, we compare appearance-based single-modal co-training with multi-modal. Our results suggest that in a standard SSL setting (no domain shift, a few human-labeled data) and under virtual-to-real domain shift (many virtual-world labeled data, no human-labeled data) multi-modal co-training outperforms single-modal. In the latter case, by performing GAN-based domain translation both co-training modalities are on pair; at least, when using an off-the-shelf depth estimation model not specifically trained on the translated images.
\end{abstract}


\blfootnote{The authors acknowledge the financial support received for this research from the Spanish TIN2017-88709-R (MINECO/AEI/FEDER, UE) project. Antonio M. López acknowledges the financial support to his general research activities given by ICREA under the ICREA Academia Program. Jose L. Gómez acknowledges the financial support to perform his PhD given by the grant FPU16/04131.}


\section{Introduction}

Supervised deep learning is enabling accurate computer vision models. Key for this success is the access to raw sensor data ({\ie}, images) with ground truth (GT) for the visual task at hand ({\eg}, image classification \cite{Sharma:2018}, object detection \cite{Ren:2015} and recognition \cite{Tang:2020}, pixel-wise instance/semantic segmentation \cite{Xie:2020, Wang:2020}, monocular depth estimation \cite{De:2021}, 3D reconstruction \cite{Kang:2020}, {\etc}). The supervised training of such computer vision models, which are based on convolutional neural networks (CNNs), is known to required very large amounts of images with GT \cite{Sun:2017}. While, until one decade ago, acquiring representative images was not easy for many computer vision applications ({\eg}, for onboard perception), nowadays, the bottleneck has shifted to the acquisition of the GT. The reason is that this GT is mainly obtained through human labeling, whose difficulty depends on the visual task. In increasing order of labeling time, we see that image classification requires image-level tags, object detection requires object bounding boxes (BBs), instance/semantic segmentation requires pixel-level instance/class silhouettes, and depth GT cannot be manually provided. Therefore, manually collecting such GT is time-consuming and does not scale as we wish. Moreover, this data labeling bottleneck may be intensified due to domain shifts among different image sensors, which could drive to per-sensor data labeling.

To address the curse of labeling, different meta-learning paradigms are being explored. In \emph{self-supervised learning} (SfSL) the idea is to train the desired models with the help of auxiliary tasks related to the main task. For instance, solving automatically generated jigsaw puzzles helps to obtain more accurate image recognition models \cite{Kolesnikov:2019}, while stereo and structure-from-motion (SfM) principles can provide self-supervision to train monocular depth estimation models \cite{Godard:2019}. In \emph{active learning} (AL) \cite{Settles:2012, Roy:2018}, there is a human---model collaborative loop, where the model proposes data labels, known as \emph{pseudo-labels}, and the human corrects them so that the model learns from the corrected labels too; thus, aiming at a progressive improvement of the model accuracy. In contrast to AL, \emph{semi-supervised learning} (SSL) \cite{Chapelle:2006, Engelen:2020} does not require human intervention. Instead, it is assumed the availability of a small set of off-the-shelf labeled data and a large set of unlabeled data, and both datasets must be used to obtain a more accurate model than if only the labeled data were used. In SfSL, the model trained with the help of the auxiliary tasks is intended to be the final model of interest. In AL and SSL, it is possible to use any model with the only purpose of self-labeling the data, {\ie}, producing the pseudo-labels, and then use labels and pseudo-labels for training the final model of interest.

\begin{figure}
\centering
\includegraphics[width=0.55\columnwidth]{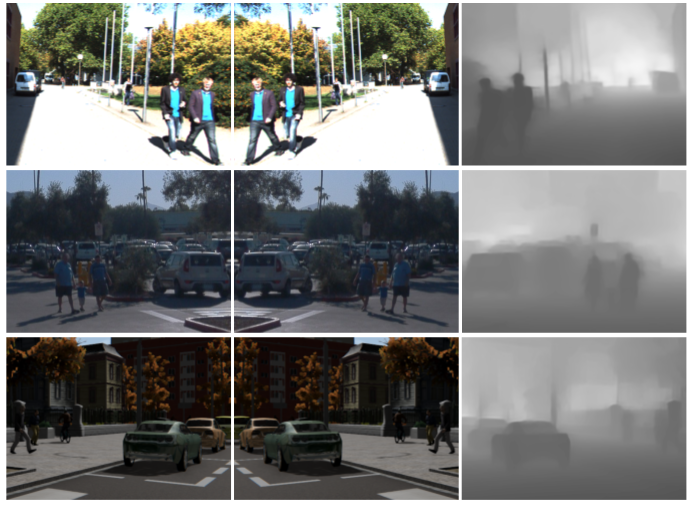}
\caption{From top to bottom: samples from KITTI ($\Kds$), Waymo ($\Wds$), and Virtual-world ($\Vds$) datasets. Middle column: cropped patch from an original image. Left column: horizontal mirror of the original patch. Right column: monocular depth estimation \cite{Yin:2019} from the original patch. Left-middle columns are the views used for co-training in \cite{Villalonga:2020}. Right-middle columns are the views also used in this paper.}
\label{fig:views}
\end{figure}

In this paper we focus on \emph{co-training} \cite{Blum:1998, Guz:2007}, a type of SSL algorithm. Co-training self-labels data through the mutual improvement of two models. These models analyze the unlabeled data according to their different \emph{views} of these data. Our work focuses on onboard vision-based perception for driver assistance and autonomous driving. In this context, vehicle and pedestrian detection are key functionalities. Accordingly, we apply co-training to significantly reduce human intervention when labeling these \emph{objects} (in computer vision terminology) for training the corresponding deep object detector. Therefore, the labels are BBs locating the mentioned traffic participants in the onboard images. More specifically, we consider two settings. On the one hand, as is usual in SSL, we assume the availability of a small set of human-labeled images ({\ie}, with BBs for the objects of interests), and a significantly larger set of unlabeled images. On the other hand, we do not assume human labeling at all, but we have a set of virtual-world images with automatically generated BBs. 

This paper is the natural continuation of the work presented by Villalonga \& L\'opez \cite{Villalonga:2020}. In this previous work, a co-training algorithm for deep object detection is presented, addressing the two above-mentioned settings too.  In \cite{Villalonga:2020}, the two views of an image consist of the original RGB representation and its horizontal mirror; thus, it is a single-modal co-training based on appearance. However, a priori, the higher difference among data views the more accurate pseudo-labels can be expected from co-training. Therefore, as a major novelty of this paper, we explore the use of two image modalities in the role of co-training views. In particular, one view is the appearance ({\ie}, the original RGB), while the other view is the corresponding depth (D) as estimated by a state-of-the-art monocular depth estimation model \cite{Yin:2019}. Thus, we term this approach as multi-modal co-training; however, it can still be considered a single-sensor because still relies only on RGB images. \Figure{views} illustrates these different views for images that we use in our experiments.

In this setting, the research questions that we address are two: (Q1) \emph{Is multi-modal (RGB/D) co-training effective on the task of providing pseudo-labeled object BBs?}; (Q2) \emph{How does perform multi-modal (RGB/D) co-training compared to single-modal (RGB)?}. After adapting the method presented in \cite{Villalonga:2020} to work with both, the single and the multi-modal data views, we ran a comprehensive set of experiments for answering these two questions. Regarding (Q1), we conclude that, indeed, multi-modal co-training is rather effective. Regarding (Q2), we conclude that in a standard SSL setting (no domain shift, a few human-labeled data) and under virtual-to-real domain shift (many virtual-world labeled data, no human-labeled data) multi-modal co-training outperforms single-modal. In the latter case, when GAN-based virtual-to-real image translation is performed \cite{Zhu:2017} ({\ie}, as image-level domain adaptation) both co-training modalities are on pair; at least, by using an off-the-shelf monocular depth estimation model not specifically trained on the translated images. 

We organize the rest of the paper as follows. Section \ref{sec:rw} reviews related works. Section \ref{sec:mm} draws the co-training algorithm. 
Section \ref{sec:er} details our experimental setting, discussing the obtained results in terms of (Q1) and (Q2). Section \ref{sec:c} summarizes the presented work, suggesting lines of continuation.

\section{Related work}
\label{sec:rw}

As we have mentioned before, co-training falls in the realm of SSL. Thus, here we summarize previous related works applying SSL methods. The input to these methods consists of a labeled dataset, $\XL$, and an unlabeled one, $\XU$, with $\#\XU\gg\#\XL$ and $\domainset{\XU} = \domainset{\XL}$, where $\#\datasets$ is the cardinality of the set $\datasets$ and $\domainset{\datasets}$ refers to the domain from which $\datasets$ has been drawn. Note that, when the latter requirement does not hold, we are under a domain shift setting. The goal of a SSL method is to use both $\XL$ and $\XU$ to allow the training of a predictive model, $\model$, so that its accuracy is higher than if only $\XL$ is used for its training. In other words, the goal is to leverage unlabeled data.

A classical SSL approach is the so-called \emph{self-training}, introduced by Yarowsky \cite{Yarowsky:1995} in the context of language processing. Self-training is an incremental process that starts by training $\model$ on $\XL$; then, $\model$ runs on $\XU$, and its predictions are used to form a pseudo-labeled set $\XP$, further used together with $\XL$ to retrain $\model$. This is repeated until convergence, and the accuracy of $\model$, as well as the quality of $\XP$, are supposed to become higher as the cycles progress. \citetal{Jeong}{Jeong:2019} used self-training for deep object detection (on PASCAL VOC and MS-COCO datasets). To collect $\XP$, a consistency loss is added while training $\model$, which is a CNN for object detection in this case, together with a mechanism for removing predominant backgrounds. The consistency loss is based on the idea that $\model(\image^{\ulab})\sim{\model({\image^{\ulab}}^\Lsh)}^\Lsh$, where $\image^{\ulab}$ is an unlabeled image, and $"\Lsh"$ refers to performing horizontal mirroring. \citetal{Lokhande}{Lokhande:2020} used self-training for deep image classification. In this case, the original activation functions of $\model$, a CNN for image classification, must be changed to Hermite polynomials. Note that these two examples of self-training involve modifications either in the architecture of $\model$ \cite{Lokhande:2020} or in its training framework \cite{Jeong:2019}. However, we aim at using a given $\model$ together with its training framework as a black box, so performing SSL only at the data level. In this way, we can always benefit from state-of-the-art models and training frameworks, {\ie}, avoiding changing the SSL approach if those change. In this way, we can also decouple the model used to produce pseudo-labels from the model that would be trained with them for deploying the application of interest.

A major challenge when using self-training is to avoid drifting to erroneous pseudo-labels. Note that, if $\XP$ is biased to some erroneous pseudo-labels, when using this set to retrain $\model$ incrementally, a point can be reached where $\XL$ cannot compensate the errors in $\XP$, and $\model$ may end learning wrong data patterns and so producing more erroneous pseudo-labels. Thus, as alternative to the self-training of Yarowsky \cite{Yarowsky:1995}, Blum and Mitchell proposed \emph{co-training} \cite{Blum:1998}. Briefly, co-training is based on two models, $\model_{v_1}$ and $\model_{v_2}$, each one incrementally trained on different data features, termed as \emph{views}. In each training cycle, $\model_{v_1}$ and $\model_{v_2}$ \emph{collaborate} to form $\XP=\XP_{v_1}\cup\XP_{v_2}$. Where, $\XP_{v_i}$ and $\XL$ are used to retrain $\model_{v_i}, i\in\{1,2\}$. This is repeated until convergence. It is assumed that each view, ${v_i}$, is discriminant enough as to train an accurate $\model_{v_i}$. Different implementations of co-training, may differ in the collaboration policy. Our approach follows the \emph{disagreement} idea introduced by \citetal{Guz}{Guz:2007} in the context of sentence segmentation, later refined by Tur \cite{Tur:2009} to address domain shifts in the context of natural language processing. In short, only pseudo-labels of high confidence for $\model_{v_i}$ but of low confidence for $\model_{v_j}$, $i,j\in\{1,2\},i \neq j$, are considered as part of $\XP_{v_j}$ in each training cycle. Soon, disagreement-based SSL attracted much interest \cite{Zhou:2010}. 
In general, $\model_{v_1}$ and $\model_{v_2}$ can be based on different data views by either training on different data samples ($\XP_{v_1}\neq\XP_{v_2}$) or being different models ({\eg}, $\model_{v_1}$ and $\model_{v_2}$ can be based on two different CNN architectures). The disagreement-based co-training falls in the former case. In this line, \citetal{Qiao}{Qiao:2018} used co-training for deep image classification, where the two different views are achieved by training on mutually adversarial samples. However, this implies linking the training of the $\model_{v_i}$'s at the level of the loss function, while, as we have mentioned before, we want to use these models as black boxes. 

\begin{figure}[!t]
\centering
     \begin{subfigure}[b]{0.49\textwidth}
         \centering
            \includegraphics[trim=0 5 0 4, clip, width=\columnwidth]{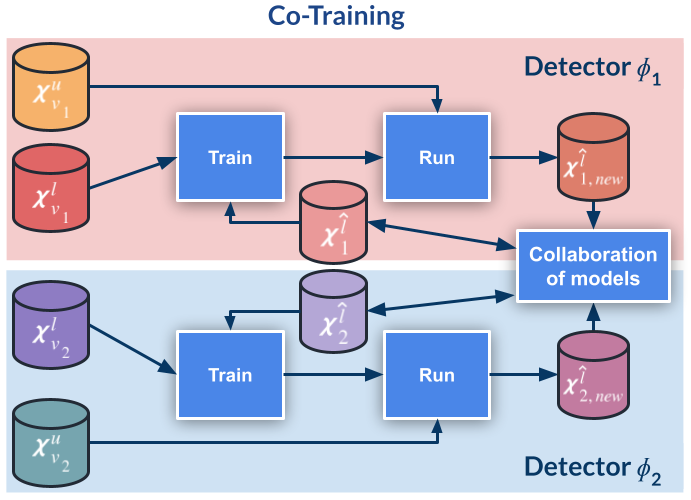}\\
     \end{subfigure}
     \hfill
     \begin{subfigure}[b]{0.48\textwidth}
         \centering
            \includegraphics[trim=0 7 0 5, clip, width=\columnwidth]{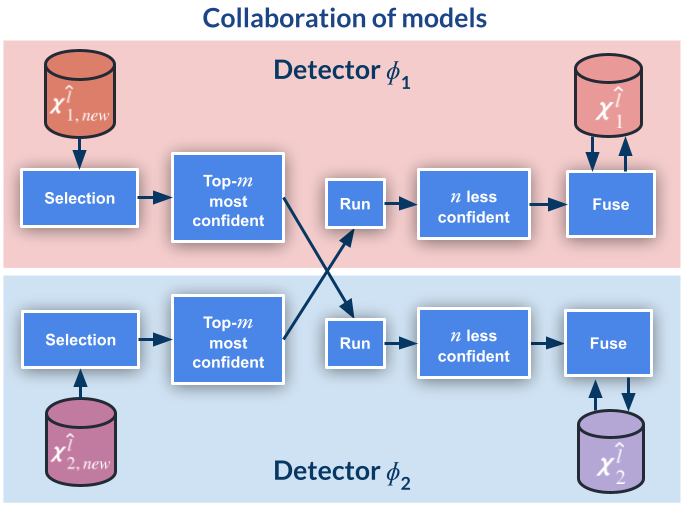}
    \end{subfigure}
\caption{Co-training pipeline: the left diagram shows the global block structure, while the right diagram details the \emph{collaboration of models} block. Symbols and procedures are based on \Alg{co-training}. We refer to this algorithm and the main text for a detailed explanation.}
\label{fig:cotrain}
\end{figure}

The most similar work to this paper is the co-training framework that we introduced in \cite{Villalonga:2020} since we work on top of it. In \cite{Villalonga:2020}, two single-modal views are considered. These consist of using $\model_{v_1}$ to process the original images from $\XU$ while using $\model_{v_2}$ to process their horizontally mirrored counterparts, and analogously for $\XL$. A disagreement-based collaboration is applied to form $\XP_{v_1}$ and $\XP_{v_2}$. Moreover, not only the setting where $\XL$ is based on human labels is considered, but also when it is based on virtual-world data. In the latter case, a GAN-based virtual-to-real image translation \cite{Zhu:2017} is used as pre-processing for the virtual-world images, {\ie}, before taking them for running the co-training procedure. Very recently, \citetal{D\'{i}az}{Diaz:2021} presented co-training for visual object recognition. In other words, the paper addresses a classification problem, while we address both localization and classification to perform object detection. While the different views proposed in \cite{Diaz:2021} rely on self-supervision ({\eg}, forcing image rotations), here, these rely on data multi-modality. In fact, in our previous work \cite{Villalonga:2020}, we used mirroring to force different data views, which can be considered as a kind of self-supervision too. Here, after adapting and improving the framework used in \cite{Villalonga:2020}, we confront this previous setting to a new multi-modal single-sensor version (\Alg{co-training} and \Fig{cotrain}). We focus on the case where $\model_{v_1}$ works with the original images while $\model_{v_2}$ works with their estimated depth. Analyzing this setting is quite interesting because appearance and depth are different views of the same data.

We need an out-of-the-shelf monocular depth estimation (MDE) model, so that we can keep the co-training as a single-sensor even being multi-modal. MDE can be based on either LiDAR supervision, or stereo/SfM self-supervision, or combinations; where, both LiDAR and stereo data, and SfM computations, are only required at training time, but not at testing time. We refer to \cite{De:2021} for a review on MDE state-of-the-art. In this paper, to isolate the multi-modal co-training performance assessment as much as possible from the MDE performance, we have chosen the top-performing supervised method proposed by \citetal{Yin}{Yin:2019}.

Finally, we would like to mention that there are methods in the literature that may be confused with co-training, so it is worth introducing a clarification note. This is the case of the \emph{co-teaching} proposed by \citetal{Han}{Han:2018} and the \emph{co-teaching+} of \citetal{Yu}{Yu:2019}. These methods have been applied to deep image classification to handle noisy labels on $\XL$. However, citing \citetal{Han}{Han:2018}, \emph{co-training is designed for SSL, and co-teaching is for learning with noisy (ground truth) labels (LNL); as LNL is not a special case of SSL, we cannot simply translate co-training from one problem setting to another problem setting}. 

\vspace{2\baselineskip}
\begin{algorithm}[!h]
\caption{Self-labeling of object BBs by co-training.}
\label{alg:co-training}
\DontPrintSemicolon
\SetAlgoLined
\SetNoFillComment
\LinesNotNumbered 
\SetKwInOut{Input}{Input}\SetKwInOut{Output}{Output}
\Input{{\small View-paired sets of labeled images:} $\XL_{v_1}, \XL_{v_2}$\\ 
       {\small View-paired sets of unlabeled images:} $\XU_{v_1}, \XU_{v_2}$\\
       {\small Object detection architecture, and its training hyper-parameters:} $\CNN, \HPOD$\\
       {\small Co-training hyper-parameters:} $\HPCoT=\{T,N,n,m,\HPStop[,\HPSeq]\}$\\}
\Output{{\small New labeled images:} $\XP\subseteq\XU_{v_1}$}
\tcp*[l]{}
\tcp*[l]{{\bf Initialize} models and working datasets.}
\tcp*[l]{}
\begin{tabularx}{\textwidth}{r>{}c>{}l}
$<\XP_1,\XP_2,k>$                & $\gets$ & $<\emptyset,\emptyset,0>$\\
$\OD_1, \OD_2$                   & $\gets$ & $\TD{\CNN}{\HPOD}{\XL_{v_1}}{\XP_1}, \TD{\CNN}{\HPOD}{\XL_{v_2}}{\XP_2}$\\
$\XP_{1,new}, \XP_{2,new}$       & $\gets$ & $\RD{\OD_1}{\XU_{v_1}}{T}, \RD{\OD_2}{\XU_{v_2}}{T}$\\
\end{tabularx}
\Repeat{$\STOP{\HPStop}{\XP_{old}}{\XP_{1,new}}{k\mbox{++}}$}{
\begin{tabularx}{\textwidth}{r>{}c>{}l}
$\XP_{old}$                              & $\gets$ & $\XP_{1,new}$\\
&&\\&& \tcp*[l]{{\bf Collaboration} of models.}\\&&\\
$\XP_{1,\uparrow}, \XP_{2,\uparrow}$     & $\gets$ & $\Select{\uparrow}{m}{\Rand{\XP_{1,new}}{N}{[,\HPSeq,k]}}, \Select{\uparrow}{m}{\Rand{\XP_{2,new}}{N}{[,\HPSeq,k]}}$\\
$\XP_{1,\downarrow}, \XP_{2,\downarrow}$ & $\gets$ & $\Select{\downarrow}{n}{\RD{\OD_1}{\XP_{2,\uparrow}}{T}}, \Select{\downarrow}{n}{\RD{\OD_2}{\XP_{1,\uparrow}}{T}}$\\
$\XP_1, \XP_2$                           & $\gets$ & $\Fuse{\XP_1}{\XP_{1,\downarrow}}, \Fuse{\XP_2}{\XP_{2,\downarrow}}$\\
&&\\&& \tcp*[l]{{\bf Retrain} models and {\bf Update} working datasets.}\\&&\\
$\OD_1, \OD_2$                           & $\gets$ & $\TD{\CNN}{\HPOD}{\XL_{v_1}}{\XP_1}, \TD{\CNN}{\HPOD}{\XL_{v_2}}{\XP_2}$\\
$\XP_{1,new}, \XP_{2,new}$               & $\gets$ & $\RD{\OD_1}{\XU_{v_1}}{T}, \RD{\OD_2}{\XU_{v_2}}{T}$\\
\end{tabularx}
}
\begin{tabularx}{\textwidth}{r>{}c>{}l}
$\XP$ & $\gets$ & $\XP_{1,new}$\\
\end{tabularx}
\KwRet{$\XP$}
\end{algorithm}

\newpage
\section{Method}
\label{sec:mm}
In this section, we explain our co-training procedure with the support of \Fig{cotrain} and \Alg{co-training}. Up to a large extent, we follow the same terminology as in \cite{Villalonga:2020}.\\ 

\begin{table}
\centering
\caption{The different configurations that we consider for \Alg{co-training} in this paper, according to the input datasets. In the single-modal cases, we work only with RGB images (appearance), either from a real-world dataset ($\Rdspar{RGB}$), or a virtual-world one ($\Vdspar{RGB}$), or a virtual-to-real domain-adapted one ($\VGRdspar{RGB}$), {\ie}, using a GAN-based $\Vdspar{RGB}\rightarrow\Rdspar{RGB}$ image translation. One view of the data ($v_1$) corresponds to the original RGB images of each set, while the other view ($v_2$) corresponds to their horizontally mirrored counterparts, indicated with the symbol $"\Lsh"$. In the multi-modal cases, view $v_1$ is the same as for the single-modal case (RGB), while view $v_2$ corresponds to the depth (D) estimated from the RGB images by using an off-the-shelf monocular depth estimation model.}\label{tab:cotraining-settings}
\begin{center}
\begin{tabular}{cccccc}
\toprule
Modality                       & Domain shift? & $\XL_{v_1}$      & $\XL_{v_2}$               & $\XU_{v_1}$ & $\XU_{v_2}$ \\
\bottomrule
\multirow{3}{*}{Single-modal}  & No            & $\Rdspar{RGB}$   & ${\Rdspar{RGB}}^{\Lsh}$   & \multicolumn{2}{c}{\multirow{3}{*}{$\Rdspar{RGB}$}}\\
                               & Yes           & $\Vdspar{RGB}$   & ${\Vdspar{RGB}}^{\Lsh}$   & \\
                               & Adapted       & $\VGRdspar{RGB}$ & ${\VGRdspar{RGB}}^{\Lsh}$ & \\
\midrule
\multirow{3}{*}{Multi-modal}   & No            & $\Rdspar{RGB}$   & ${\Rdspar{D}}$            & \multirow{3}{*}{$\Rdspar{RGB}$} & \multirow{3}{*}{$\Rdspar{D}$}\\
                               & Yes           & $\Vdspar{RGB}$   & ${\Vdspar{D}}$            & \\
                               & Adapted       & $\VGRdspar{RGB}$ & ${\VGRdspar{D}}$          & \\
\bottomrule
\end{tabular}

\end{center}
\end{table}

\noindent\textbf{Input:} The specific sets of labeled ($\XL_{v_1}, \XL_{v_2}$) and unlabeled ($\XU_{v_1}, \XU_{v_2}$) input data in \Alg{co-training} determine if we are running on either a single or multi-modal setting. Also, if we are supported or not by virtual-world images or their virtual-to-real translated counterparts. \Tab{cotraining-settings}, clarifies the different co-training settings depending on these datasets. In \Alg{co-training}, \emph{view-paired sets} means that each image of one set has a counterpart in the other, {\ie}, following \Tab{cotraining-settings}, its horizontal mirror or its estimated depth. Since the co-training is agnostic to the specific object detector in use, we explicitly consider its corresponding CNN architecture, $\CNN$, and training hyper-parameter, $\HPOD$, as inputs. Finally, $\HPCoT$ consists of the co-training hyper-parameters, which we will introduce while explaining the part of the algorithm in which each of them is required.\\

\noindent\textbf{Output:} It consists in a set of images ($\XP$) from $\XU_{v_1}$, for which co-training is providing pseudo-labels, {\ie}, object BBs in this paper. In our experiments, according to \Tab{cotraining-settings}, $\XU_{v_1}$ always corresponds to the unlabeled set of original real-world images. Since we consider as output a set of self-labeled images, which complement the input set of labeled images, they can be later used to train a model based on $\CNN$ or any other CNN architecture performing the same task ({\ie}, requiring the same type of BBs).\\

\noindent\textbf{Initialize:} First, the initial object detection models ($\OD_1, \OD_2$) are trained using the respective views of the labeled data ($\XL_{v_1}, \XL_{v_2}$). After their training, these models are applied to the respective views of the unlabeled data ($\XU_{v_1}, \XU_{v_2}$). Detections ({\ie}, object BBs) with a confidence over a threshold are considered pseudo-labels. Since we address a multi-class problem, per-class thresholds are contained in the set $T$, a hyper-parameter in $\HPCoT$. The temporary self-labeled sets generated by $\OD_1$ and $\OD_2$ are $\XP_{1,new}$ and $\XP_{2,new}$, respectively. At this point no collaboration is produced between $\OD_1$ and $\OD_2$. In fact, while co-training loops (repeat body), the self-labeled sets resulting from the collaboration are $\XP_1$ and $\XP_2$, which are initialized as empty. In the training function, $\TD{\CNN}{\HPOD}{\xL}{{\xP}}:\OD$, we use BB labels (in $\xL$) and BB pseudo-labels (in $\xP$) indistinctly. However, we only consider background samples from $\xL$, since, as co-training progresses, $\xP$ may be instantiated with a set of self-labeled images containing false negatives ({\ie}, undetected objects) which could be erroneously taken as hard negatives ({\ie}, background quite similar to objects) when training $\OD$.\\

\noindent\textbf{Collaboration:} The two object detection models collaborate by exchanging pseudo-labeled images (\Fig{cotrain}-right). This exchange is inspired in disagreement-based SSL \cite{Zhou:2010}. Our specific approach is controlled by the co-training hyper-parameters $N, n, m$, and, in case of working with image sequences instead of with sets of isolated images, also by $\HPSeq=\{{\Delta t}_1,{\Delta t}_2\}, {\Delta t}_1, {\Delta t}_2$. This approach consists of the following three steps.

\textbf{\emph{1st Step)}} Each model selects the set of its {top-$m$} most confident self-labeled images ($\XP_{1,\uparrow}, \XP_{2,\uparrow}$); where, the confidence of an image is defined as the average over the confidences of the pseudo-labels of the image, {\ie}, in our case, over the object detections. Thus, $\XP_{i,\uparrow}\subseteq\XP_{i,new}, i\in\{1,2\}$. However, for creating $\XP_{i,\uparrow}$, we do not consider all the self-labeled images in $\XP_{i,new}$. Instead, to minimize bias and favor speed, we only consider $N$ randomly selected images from $\XP_{i,new}$. In the case of working with image sequences, to favor variability in the pseudo-labels, the random choice is constrained to avoid using consecutive frames. This is controlled by thresholds ${\Delta t}_1$ and ${\Delta t}_2$; where ${\Delta t}_1$ controls the minimum frame distance between frames selected at the current co-training cycle ($k$), and ${\Delta t}_2$ among frames at current cycle with respect to frames selected in previous cycles ($<k$). We apply ${\Delta t}_1$ first, then ${\Delta t}_2$, and then the random selection among the frames passing these constraints. 

\textbf{\emph{2nd Step)}} Model $\OD_i$ processes $\XP_{j,\uparrow}, i,j\in\{1,2\}, i\neq j$, keeping the set of the $n$ less confident self-labeled images for it. Thus, we obtain the new sets $\XP_{1,\downarrow}$ and $\XP_{2,\downarrow}$. Therefore, considering the first and second steps, we see that one model shares with the other those images that it has self-labeled with more confidence, and, of these, each model retains for retraining those that it self-labels with less confidence. Therefore, this step implements the actual collaboration between models $\OD_1$ and $\OD_2$.

\textbf{\emph{3rd Step)}} The self-labeled sets obtained in previous step ($\XP_{1,\downarrow}, \XP_{2,\downarrow}$) are fused with those accumulated from previous co-training cycles ($\XP_1, \XP_2$). This is done by properly calling the function $\Fuse{\xP_{old}}{\xP_{new}}:\xP$ for each view. The returned set of self-labeled images, $\xP$, contains $\xP_{old}\cup\xP_{new} - \xP_{old}\cap\xP_{new}$, and, from $\xP_{old}\cap\xP_{new}$, only those self-labeled images in $\xP_{new}$ are added to $\xP$.\\ 

\noindent\textbf{Retrain \& Update:} At this point we have new sets of self-labeled images ($\XP_1, \XP_2$), which, together with the corresponding input labeled sets ($\XL_{v_1}, \XL_{v_2}$), are used to retrain the models $\OD_1$ and $\OD_2$. Afterwards, these new models are used to obtain new temporary self-labeled set ($\XP_{1,new}, \XP_{2,new}$) through their application to the corresponding unlabeled sets ($\XU_{v_1}, \XU_{v_2}$). Then, co-training can start a new cycle.\\

\noindent\textbf{Stop:} The function {$\STOP{\HPStop}{\xP_{old}}{\xP_{new}}{k}:{\small\mbox{Boolean}}$} determines if a new co-training cycle is executed. This is controlled by the co-training hyper-parameters $\HPStop=\{K_{min},K_{max},T_{\Delta_{mAP}},{\Delta K}\}$. Co-training will execute a minimum of $K_{min}$ cycles and a maximum of $K_{max}$, being $k$ the current number. The parameters $\xP_{old}$ and $\xP_{new}$ are supposed to be instantiated with the sets of self-labeled images in previous and current co-training cycles, respectively. The similarity of these sets is monitored in each cycle, so that if its stable for more than $\Delta K$ consecutive cycles, convergence is assumed and co-trained stopped. This constrain could already be satisfied at $k=K_{min}$ provided $K_{min}\geq{\Delta K}$. The metric used to compute the similarity between these self-labeled sets is mAP (mean average precision) \cite{Geiger:2012}, where $\xP_{old}$ plays the role of GT and $\xP_{new}$ the role of results under evaluation. Then, mAP is considered stable between two consecutive cycles if its magnitude variation is below the threshold $T_{\Delta_{mAP}}$.

\section{Experimental results}
\label{sec:er}

\subsection{Datasets and evaluation protocol}
We follow the experimental setup of \cite{Villalonga:2020}. Therefore, we use KITTI \cite{Geiger:2012} and Waymo \cite{Sun:2020} as real-world datasets, here denoted as $\Kds$ and $\Wds$, respectively. We use a variant of the SYNTHIA dataset \cite{Ros:2016} as virtual-world data, here denoted as $\Vds$. For $\Kds$ we use \citetal{Xiang}{Xiang:2015} split, which reduces the correlation between training and testing data. While this implies that $\Kds$ is formed by isolated images, $\Wds$ is composed of image sequences. To align its acquisition conditions with $\Kds$, we consider daytime sequences without adverse weather. From them, as recommended in \cite{Sun:2020}, we randomly select some sequences for training and the rest for testing. Furthermore, we adapt {$\Wds$'s} image size to match $\Kds$ (i.e. $1240\times375$ pixels) by first eliminating the top rows of each image so avoiding large sky areas, and then selecting a centered area of $1240$ pixel width. The 2D BBs of $\Wds$ and $\Vds$, are obtained by projecting the available 3D BBs. On the other hand, $\Vds$ is generated by mimicking some acquisition conditions of $\Kds$, such as image resolution, non-adverse weather, daytime, and only considering isolated shots instead of image sequences. Besides, {$\Vds$'s images} include standard visual post-effects such as anti-aliasing, ambient occlusion, depth of field, eye adaptation, blooming, and chromatic aberration. In the following, we term as $\Kdstrain$ and $\Kdstest$ the training and testing sets of $\Kds$, respectively. Analogously,  $\Wdstrain$ and $\Wdstest$ are the training and testing sets of $\Wds$. For each dataset, Table \ref{tab:datasets} summarizes the number of images and object BBs (vehicles and pedestrians) used for training and testing our object detectors. Note that $\Vds$ is only used for training purposes. 

\begin{table}
\centering
\caption{Datasets ($\datasets$): train ($\Xtrain$) and test ($\Xtest$) statistics, $\datasets=\Xtrain\cup\Xtest, \Xtrain\cap\Xtest=\emptyset$.}\label{tab:datasets}

\begin{tabular}{@{}lcccccc@{}}
\toprule
                       & \multicolumn{3}{c}{$\Xtrain$}   & \multicolumn{3}{c}{$\Xtest$}    \\
Dataset ($\datasets$)  & Images & Vehicles & Pedestrians & Images & Vehicles & Pedestrians \\
\toprule
Virtual ($\Vds$)       & 19,791 & 43,326   & 44,863      &        &          &             \\
\midrule
KITTI ($\Kds$)         &  3,682 & 14,941   &  3,154      & 3,799  & 18,194   & 1,333      \\
\midrule
Waymo ($\Wds$)         &  9,873 & 64,446   &  9,918      & 4,161  & 24,600   & 3,068       \\
\bottomrule
\end{tabular}

\end{table}

We apply the KITTI benchmark protocol for object detection \cite{Geiger:2012}. Furthermore, following \cite{Villalonga:2020}, we focus on the so-called moderate difficulty, which implies that the minimum BB height to detect objects is $25$ pixels for $\Kds$ and $50$ pixels for $\Wds$. Once co-training finishes, we use the labeled data ($\XL$) and the data self-labeled by co-training ($\XP$) to train the final object detector, namely, $\OD_F$. Since this is the ultimate goal, we use the accuracy of such a detector as metric to evaluate the effectiveness of the co-training procedure. If it performs well at self-labeling objects, the accuracy of $\OD_F$ should be close to the upper-bound ({\ie}, when the 100\% of the real-world labeled data used to train $\OD_F$ is provided by humans), otherwise, the accuracy of $\OD_F$ is expected to be close to the lower-bound ({\ie}, when using either a small percentage of human-labeled data or only virtual-world data to train $\OD_F$). 

\subsection{Implementation details}

When using virtual-world images we not only experiment with the originals but also with their GAN-based virtual-to-real translated counterparts, {\ie}, aiming at closing the domain shift between virtual and real worlds. Since the translated images are the same for both co-training modalities, we take them from \cite{Villalonga:2020}, where a CycleGAN \cite{Zhu:2017} was used to learn the translations $\GANKds:\Vds\rightarrow\Kds$ and $\GANWds:\Vds\rightarrow\Wds$. To obtain these images, CycleGAN training was done for 40 epochs using a weight of 1.0 for the identity mapping loss, and a patch-wise strategy with patches of 300x300 pixels, while keeping the rest of the parameters as recommended in \cite{Zhu:2017}. We denote as $\VGKds=\GANKds(\Vds)$ and $\VGWds=\GANWds(\Vds)$ the sets of virtual-world images transformed by $\GANKds$ and $\GANWds$, respectively. 
The 2D BBs in $\Vds$ are used for $\VGKds$ and $\VGWds$. Furthermore, note that analogously to $\Vds$, $\VGKds$ and $\VGWds$ are only used for training. For multi-modal co-training, depth estimation is applied indistinctly to the real-world datasets, the virtual-world one, and the GAN-based translated ones.

\begin{table}
\centering
\caption{Co-training hyper-parameters as defined in \Alg{co-training}. We use the same values for $\Kds$ and $\Wds$ datasets, but $\HPSeq$ only applies to $\Wds$. $N$, $n$, $m$, ${\Delta t}_1$, and ${\Delta t}_2$ are set in number-of-images units, $K_{min}, K_{max}$ and ${\Delta K}$ in number-of-cycles, $T_{\Delta_{mAP}}$ runs in [0..100]. $T$ hyper-parameter contains the confidence detection thresholds for vehicles and pedestrians, which run in $[0..1]$, and we have set the same value for both. The setting $m=\infty$ means that all the images self-labeled at current co-training cycle are exchanged by the models $\OD_1$ and $\OD_2$ for collaboration, {\ie}, these will then  select the $n$ less confident for them.}\label{tab:parameters}

\begin{tabular}{@{}ccccccccccccc@{}}
\toprule
            &       &     &              && \multicolumn{4}{c}{$\HPStop$}                             && \multicolumn{2}{c}{$\HPSeq$} \\
$T$         & $N$   & $n$ & $m$          && $K_{min}$ & $K_{max}$ & ${\Delta K}$ & $T_{\Delta_{mAP}}$ && ${\Delta t}_1$ & ${\Delta t}_2$ \\
\midrule
\{0.8,0.8\} & 500   & 100 & $\infty$     && 20        & 30        & 5            & 2.0                && 5              & 10 \\
\bottomrule
\end{tabular}
\end{table}

\begin{figure*}
\centering
\includegraphics[width=1\textwidth]{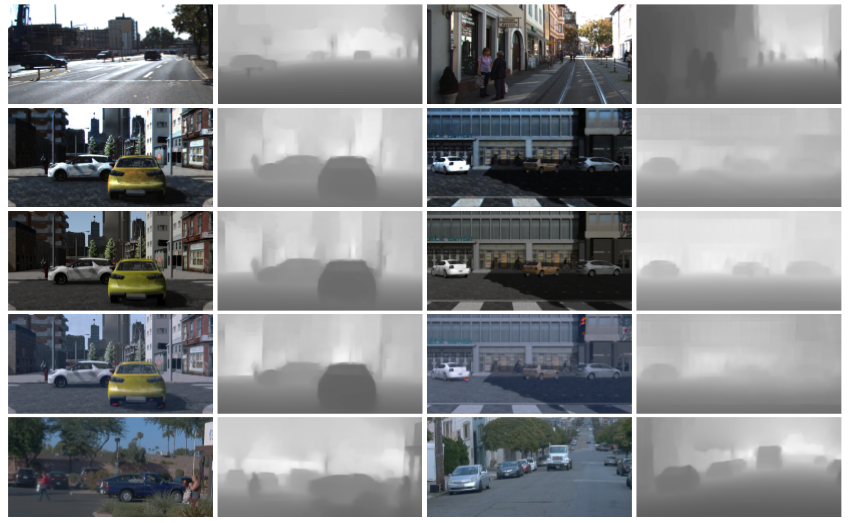}
\caption{RGB images with their estimated depth. From top to bottom rows: samples from $\Kds$, $\VGKds$, $\Vds$, $\VGWds$, $\Wds$. The samples of $\VGKds$ and $\VGWds$ correspond to transforming the samples of $\Vds$ to $\Kds$ and $\Wds$ domains, respectively. The monocular depth estimation model \cite{Yin:2019} was trained on the $\Kds$ domain.} 
\label{fig:monodepth}
\end{figure*}

In the multi-modal setting, one of the co-training views is the appearance (RGB) and the other is the corresponding estimated depth (D). To keep co-training single-sensor, we use monocular depth estimation (MDE). In particular, we leverage a state-of-the-art MDE model publicly released by \citetal{Yin}{Yin:2019}. It has been trained on KITTI data, thus, being ideal to work with $\Kds$. However, since our aim is not to obtain accurate depth estimation, but to generate an alternative data view useful to detect the objects of interest, we have used the same MDE model for all the considered datasets. Despite this, \Fig{monodepth} shows how the estimated depth properly captures the depth structure for the images of all datasets, {\ie}, not only for $\Kds$, but also for $\Wds, \Vds, \VGKds$ and $\VGWds$. However, we observe that the depth structure for {$\VGKds$'s} and {$\VGWds$'s} images is more blurred at far distances than for $\Vds$, especially for $\VGWds$.

Following \cite{Villalonga:2020}, we use Faster R-CNN with a VGG16 feature extractor (backbone) as the CNN architecture for object detection, {\ie}, as $\CNN$ in \Alg{co-training}. In particular, we rely on the \emph{Detectron} implementation \cite{Detectron:2018}. For training, we always initialize VGG16 with ImageNet pre-trained weights, while the weights of the rest of the CNN ({\ie}, the candidates' generator and classifier stages) are randomly initialized. Faster R-CNN training is based on 40,000 iterations of the SGD optimizer. Note that these iterations refer to the function $\TD{\CNN}{\HPOD}{\xL}{{\xP}}:\OD$ in \Alg{co-training}, not to co-training cycles. Each iteration uses a mini-batch of two images randomly sampled from $\xL\cup\xP$. Thus, looking at how $\TD{\CNN}{\HPOD}{\xL}{{\xP}}:\OD$ is called in \Alg{co-training}, we can see that, for each view, the parameter $\xL$ receives the same input in all co-training cycles, while $\xP$ changes from cycle-to-cycle. The SGD learning rate starts at 0.001 and we set a decay of 0.1 at iterations 30,000 and 35,000. In the case of multi-modal co-training, we use horizontal mirroring as a data augmentation technique. However, we cannot do it in the case of single-modal co-training because both data views would highly correlate. Note that, as it was done in \cite{Villalonga:2020} and we can see in \Tab{cotraining-settings}, horizontal mirroring is the technique used to generate one of the data views in single-modal co-training. In terms of \Alg{co-training}, all these settings are part of $\HPOD$ and they are the same to train both $\OD_1$ and $\OD_2$. The values set for the co-training hyper-parameters are shown in \Tab{parameters}.

Finally, note that the final detection model used for evaluations, $\OD_F$, could be based on any CNN architecture for object detection, provided the GT it expects consists of 2D BBs. However, for the sake of simplicity, we also rely on Faster R-CNN to obtain $\OD_F$.

\subsection{Results}
\label{ssec:results}

To include multi-modality we improved and adapted the code used in \cite{Villalonga:2020}. For this reason, we not only execute the multi-modal co-training experiments but also redo the single-modal and baseline ones. The conclusions in \cite{Villalonga:2020} remain, but by repeating these experiments, all the results presented in this paper are based on the same code.

\begin{table}
\centering
\caption{SSL (co-training) results on vehicle (V) and pedestrian (P) detection, reporting mAP. From a training set $\Xtrain\in\{\Kdstrain, \Wdstrain\}$, we preserve the labeling information for a randomly chosen {$p$\%} of its images, while it is ignored for the rest. We report results for $p$=100 (all labels are used), $p$=5 and $p$=10. If $\Xtest=\Kdstest$, then $\Xtrain=\Kdstrain$; analogously, when $\Xtest=\Wdstest$, then $\Xtrain=\Wdstrain$, {\ie}, there is no domain shift in these experiments. Co-T (RGB) and Co-T (RGB/D) stand for single and multi modal co-training, respectively. UP and LB stand for upper bound and lower bound, respectively. Bold results indicate \textbf{best performing} within the block, where blocks are delimited by horizontal lines. \textul{Second best} is underlined, but if the difference with the best is below 0.5 points, we use bold too. $\Delta\{\OD_{F_1}~\mbox{{\vs}}~\OD_{F_2}\}$ stands for mAP of $\OD_{F_1}$ minus mAP of $\OD_{F_2}$.}\label{tab:sslresults}

\begin{tabular}{@{}lcccccc@{}}
\toprule
&\multicolumn{3}{c}{$\Xtest=\Kdstest$} & \multicolumn{3}{c}{$\Xtest=\Wdstest$} \\
Training set               & V & P & V\&P & V & P & V\&P  \\
\toprule
100\% Labeled (RGB) / UB   & \textbf{83.43} & \textbf{67.77} & \textbf{75.60} & \textbf{61.71} & \textbf{57.74} & \textbf{59.73} \\
100\% Labeled (D) / UB     & 80.80          & 53.43          & 67.12          & 55.14          & 37.67          & 46.41 \\
\midrule
5\% Labeled (RGB) / LB     & 65.20          & 46.08          & 55.64          & 51.69          & 41.92          & 46.81 \\
5\% Labeled (D) / LB       & 64.45          & 26.70          & 45.58          & 45.21          & 29.98          & 36.70 \\                           
5\% Labeled + Co-T (RGB)   & \textul{74.26} & \textul{55.41} & \textul{64.84} & \textul{54.00} & \textul{56.34} & \textul{55.17} \\
5\% Labeled + Co-T (RGB/D) & \textbf{78.64} & \textbf{57.40} & \textbf{68.02} & \textbf{58.42} & \textbf{56.98} & \textbf{57.70} \\
\midrule
10\% Labeled (RGB) / LB    & 72.31          & 45.51          & 58.91          & 49.53          & 49.83          & 49.68 \\
10\% Labeled (D) / LB      & 69.54          & 46.31          & 57.93          & 47.93          & 33.98          & 40.96 \\                           
10\% Labeled + Co-T (RGB)  & \textul{78.63} & \textbf{60.99} & \textbf{69.81} & \textul{56.15} & \textbf{60.20} & \textbf{58.18} \\
10\% Labeled + Co-T (RGB/D)& \textbf{79.68} & \textbf{60.55} & \textbf{70.12} & \textbf{59.54} & \textul{57.17} & \textbf{58.36} \\
\midrule
$\Delta\{\mbox{{\footnotesize (5\% L. + Co-T (RGB/D)) {\vs} (5\% L. (RGB) / LB)}}\}$        
                                            & +13.44          & +11.32           & +12.38          & +6.73         & +15.06          & +10.89         \\
$\Delta\{\mbox{{\footnotesize (5\% L. + Co-T (RGB/D)) {\vs} (100\% L. (RGB) / UB)}}\}$        
                                            & -4.79          & -10.37           & -7.58          & -3.29         & -0.76          & -2.03         \\
$\Delta\{\mbox{{\footnotesize (10\% L. + Co-T (RGB/D)) {\vs} (10\% L. (RGB) / LB)}}\}$        
                                            & +7.37          & +15.04           & +11.21          & +10.01         & +7.34          & +8.68         \\
$\Delta\{\mbox{{\footnotesize (10\% L. + Co-T (RGB/D)) {\vs} (100\% L. (RGB) / UB)}}\}$        
                                            & -3.75          & -7.22           & -5.48          & -2.17         & -0.57          & -1.37         \\
\bottomrule
\end{tabular}

\end{table}

\subsubsection{Standard SSL setting}
\label{sssec:results-ssl}

We start the evaluation of co-training in a standard SSL setting, {\ie}, working only with either the $\Kds$ or $\Wds$ dataset to avoid domain shift. In this setting, the cardinality of the unlabeled dataset is supposed to be significantly higher than the cardinality of the labeled, we divide the corresponding training sets accordingly. In particular, for $\Xtrain\in\{\Kdstrain, \Wdstrain\}$, we use the {$p$\%} of $\Xtrain$ as the labeled training set ($\XL$) and the rest as the unlabeled training set ($\XU$). We explore {$p=5$} and {$p=10$}, where the corresponding $\Xtrain$ is sampled randomly once and frozen for all the experiments. \Tab{sslresults} shows the obtained results for both co-training modalities. We also report upper-bound (UB) and lower-bound (LB) results. The UB corresponds to the case {$p=100$}, {\ie}, all the BBs are human-labeled. The LBs correspond to the {$p=5$} and {$p=10$} cases without using co-training, thus, not leveraging the unlabeled data. Although in this paper we assume that $\OD_F$ will be based on RGB data alone, since we use depth estimation for multi-modal co-training, as a reference we also report the UB and LB results obtained by using the estimated depth alone to train the corresponding $\OD_F$. 

Analyzing \Tab{sslresults}, we confirm that the UB and LBs based only on the estimated depth (D) show a reasonable accuracy, although not at the level of appearance (RGB) alone. This is required for the co-training to have the chance to perform well. Aside from this, we see how, indeed, both co-training modalities clearly outperform LBs. In the {$p=5$} case, multi-modal co-training clearly outperforms single-modal in all classes (V and P) and datasets ($\Kds$ and $\Wds$). Moreover, the accuracy improvement over the LBs is significantly larger than the remaining distance to the UBs. In the {$p=10$} case, both co-training modalities perform similarly. On the other hand, for $\Kds$, the accuracy of multi-modal co-training with {$p=5$} is just {$\sim2$} points below the single-modal with {$p=10$}, and less than $1$ point for $\Wds$. Therefore, for 2D object detection, we recommend multi-modal co-training for a standard SSL setting with a low ratio of labeled {\vs} unlabeled images.

\begin{table}
\centering
\caption{SSL (co-training) results on vehicle (V) and pedestrian (P) detection, under domain shift, reported as mAP. $\XL$ refers to the human-labeled target-domain training set; thus, if $\Xtest=\Kdstest$, then $\XL=\Kdstrain$, and if $\Xtest=\Wdstest$, then $\XL=\Wdstrain$. $\XP$ consists of the same images as $\XL$, but self-labeled by co-training. Co-T (RGB), Co-T (RGB/D), UP, LB, $\Delta\{\OD_{F_1}~\mbox{{\vs}}~\OD_{F_2}\}$, bold and underlined numbers are analogous to those in \Tab{sslresults}.}
\label{tab:Synth}

\begin{tabular}{@{}lcccccc@{}}
\toprule
&\multicolumn{3}{c}{$\Xtest=\Kdstest$} & \multicolumn{3}{c}{$\Xtest=\Wdstest$} \\
Training set & V & P & V\&P & V & P & V\&P  \\
\toprule
Source ($\Vds$) / LB                        & 67.46          & 65.18          & 66.32          & 38.88          & 53.37          & 46.13 \\
Target ($\XL$)                              & \textul{83.43} & \textul{67.77} & \textul{75.60} & \textbf{61.71} & \textul{57.74} & \textul{59.73} \\
Target + Source ($\XL \& \Vds$) / UB        & \textbf{87.15} & \textbf{74.69} & \textbf{80.92} & \textul{59.97} & \textbf{62.86} & \textbf{61.42} \\
\midrule
Co-T (RGB) + Source  ($\XP \& \Vds$)        & 77.97          & \textbf{71.32} & 74.65          & 48.56          & 56.33          & 52.45 \\
Co-T (RGB/D) + Source  ($\XP \& \Vds$)      & \textbf{82.90} & 67.36          & \textbf{75.13} & \textbf{64.40} & \textbf{59.17} & \textbf{61.79} \\
\midrule
$\Delta\{\mbox{(Co-T (RGB/D) + Source) {\vs} LB}\}$        
                                            & +15.44         & +2.18          & +8.81          & +25.52         & +5.80          & +15.66         \\
$\Delta\{\mbox{(Co-T (RGB/D) + ASource ) {\vs} UB}\}$
                                            & -4.25          & -7.33          & -5.79          & -0.16          & -4.27          & -2.21         \\
\bottomrule
\end{tabular}

\end{table}

\begin{table}
\centering
\caption{SSL (co-training) results on vehicle (V) and pedestrian (P) detection, after GAN-based virtual-to-real image translation, reported as mAP. ASource (\emph{adapted source}) refers to $\VGds\in\{\VGKds,\VGWds\}$. $\XL$, $\XP$, Source, Co-T (RGB), Co-T (RGB/D), UP, LB, $\Delta\{\OD_{F_1}~\mbox{{\vs}}~\OD_{F_2}\}$, bold and underlined numbers are analogous to those in \Tab{Synth}.}
\label{tab:Synth2Real}

\begin{tabular}{@{}lcccccc@{}}
\toprule
&\multicolumn{3}{c}{$\Xtest=\Kdstest$} & \multicolumn{3}{c}{$\Xtest=\Wdstest$} \\
Training set & V & P & V\&P & V & P & V\&P  \\
\toprule
ASource ($\VGds$) / LB                      & 78.41          & 65.39          & 71.90          & 52.60          & 56.36          & 54.48 \\
Target ($\XL$)                              & \textul{83.43} & \textul{67.77} & \textul{75.60} & \textul{61.71} & \textul{57.74} & \textul{59.73} \\
Target + ASource ($\XL \& \VGds$) / UB      & \textbf{86.82} & \textbf{71.59} & \textbf{79.21} & \textbf{64.56} & \textbf{63.44} & \textbf{64.00} \\
\midrule
Co-T (RGB) + ASource  ($\XP \& \VGds$)      & \textbf{85.17} & \textbf{69.93} & \textbf{77.55} & \textbf{61.49} & \textbf{59.33} & \textbf{60.41} \\
Co-T (RGB/D) + ASource ($\XP \& \VGds$)     & 83.68          & \textbf{69.48} & 76.58          & \textbf{61.49} & 58.49          & \textbf{59.99} \\
\midrule
$\Delta\{\mbox{(Co-T (RGB) + ASource) {\vs} LB}\}$
                                            & +6.76          & +4.54          & +5.65          & +8.89          & +2.97          & +5.93         \\
$\Delta\{\mbox{(Co-T (RGB/D) + ASource) {\vs} LB}\}$
                                            & +5.27          & +4.09          & +4.68          & +8.89          & +2.13          & +5.51         \\
$\Delta\{\mbox{(Co-T (RGB) + ASource) {\vs} UB}\}$
                                            & -1.65          & -1.66          &	-1.66	       & -3.07	        & -4.11	         & -3.59        \\
$\Delta\{\mbox{(Co-T (RGB/D) + ASource) {\vs} UB}\}$
                                            & -3.14          & -2.11          & -2.63          & -3.07          & -4.95          & -4.01         \\
\bottomrule
\end{tabular}

\end{table}

\subsubsection{SSL under domain shift}
\label{sssec:results-ssl-ds}

\Tab{Synth} shows the LB results for a $\OD_F$ fully trained on virtual-world images (source domain); the results of training only on the real-world images (target domain), where these images are 100\% human-labeled ({\ie}, 100\% Labeled RGB in \Tab{sslresults}); and the combination of both, which turns out to be the UB. In the case of testing on $\Wdstest$ and having $\Vds$ involved in the training, we need to accommodate the different labeling style (mainly the margin between BBs and objects) of $\Wdstest$ and $\Vds$. This is only needed for a fair quantitative evaluation, thus, for performing such evaluation the detected BBs are resized by per-class constant factors. However, the qualitative results presented in the rest of the paper are shown directly as they come by applying the corresponding $\OD_F$, {\ie}, without applying any resizing. On the other hand, this resizing is not needed for $\Kdstest$ since its labeling style is similar enough to $\Vds$.   

According to \Tab{Synth}, both co-training modalities significantly outperform the LB. Again, multi-modal co-training outperforms single-modal, especially on vehicles. Comparing multi-modal co-training with the LB, we see improvements of ${\sim15}$ points for vehicles in $\Kds$, and ${\sim25}$ in $\Wds$. Considering the joint improvement for vehicles and pedestrians we see ${\sim8}$ points for $\Kds$, and ${\sim15}$ for $\Wds$, while the distances to the UB are of ${\sim5}$ points for $\Kds$, and ${\sim2}$ for $\Wds$. Therefore, for 2D object detection, we recommend multi-modal co-training for an SSL scenario where the labeled data comes from a virtual world, {\ie}, when no human labeling is required at all, but there is a virtual-to-real domain shift.

\begin{figure}
\centering
     \begin{subfigure}[b]{0.497\textwidth}
         \centering
            \includegraphics[trim=0 5 0 7, clip, width=\columnwidth]{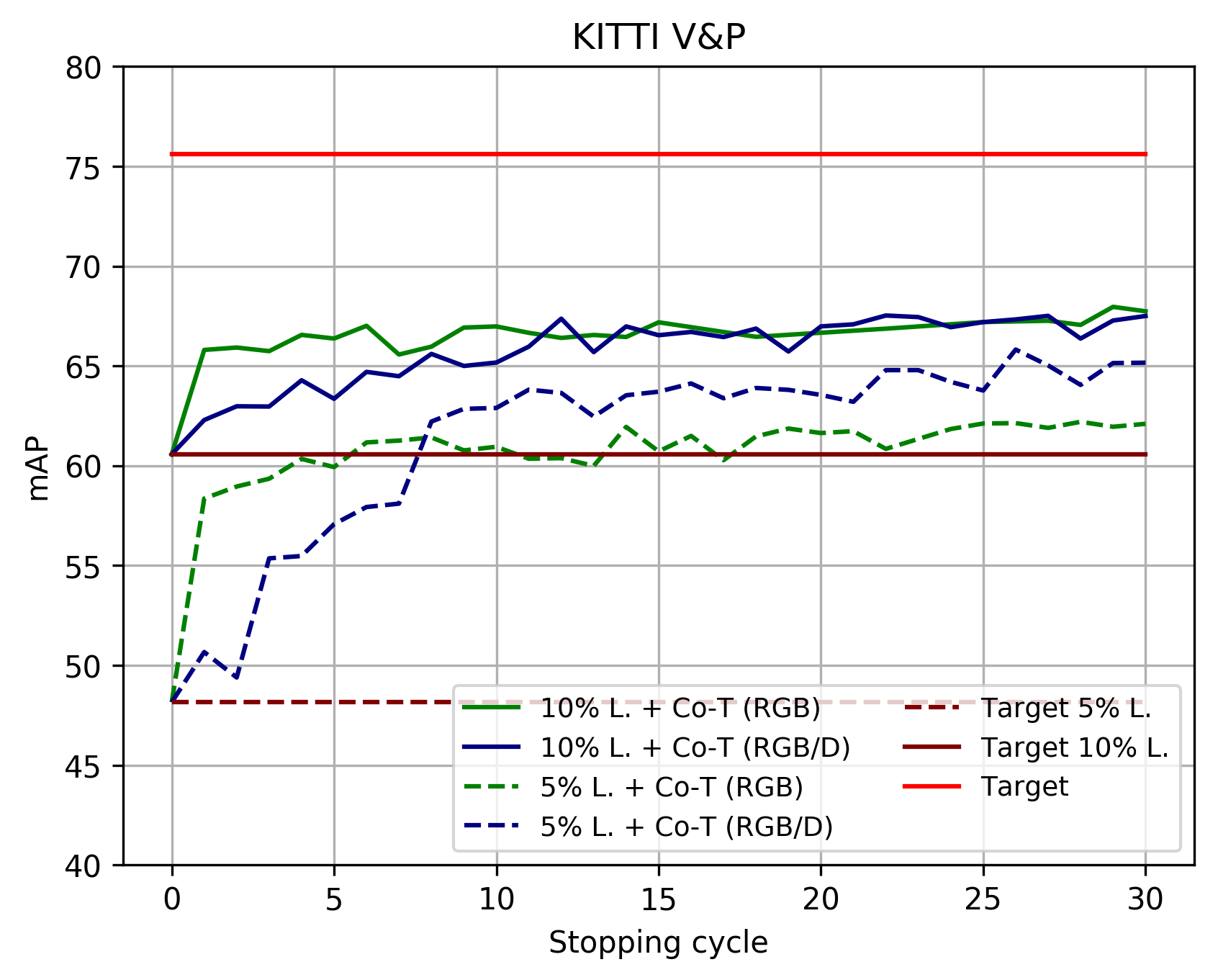}\\
     \end{subfigure}
     \hfill
     \begin{subfigure}[b]{0.497\textwidth}
         \centering
            \includegraphics[trim=0 7 0 5, clip, width=\columnwidth]{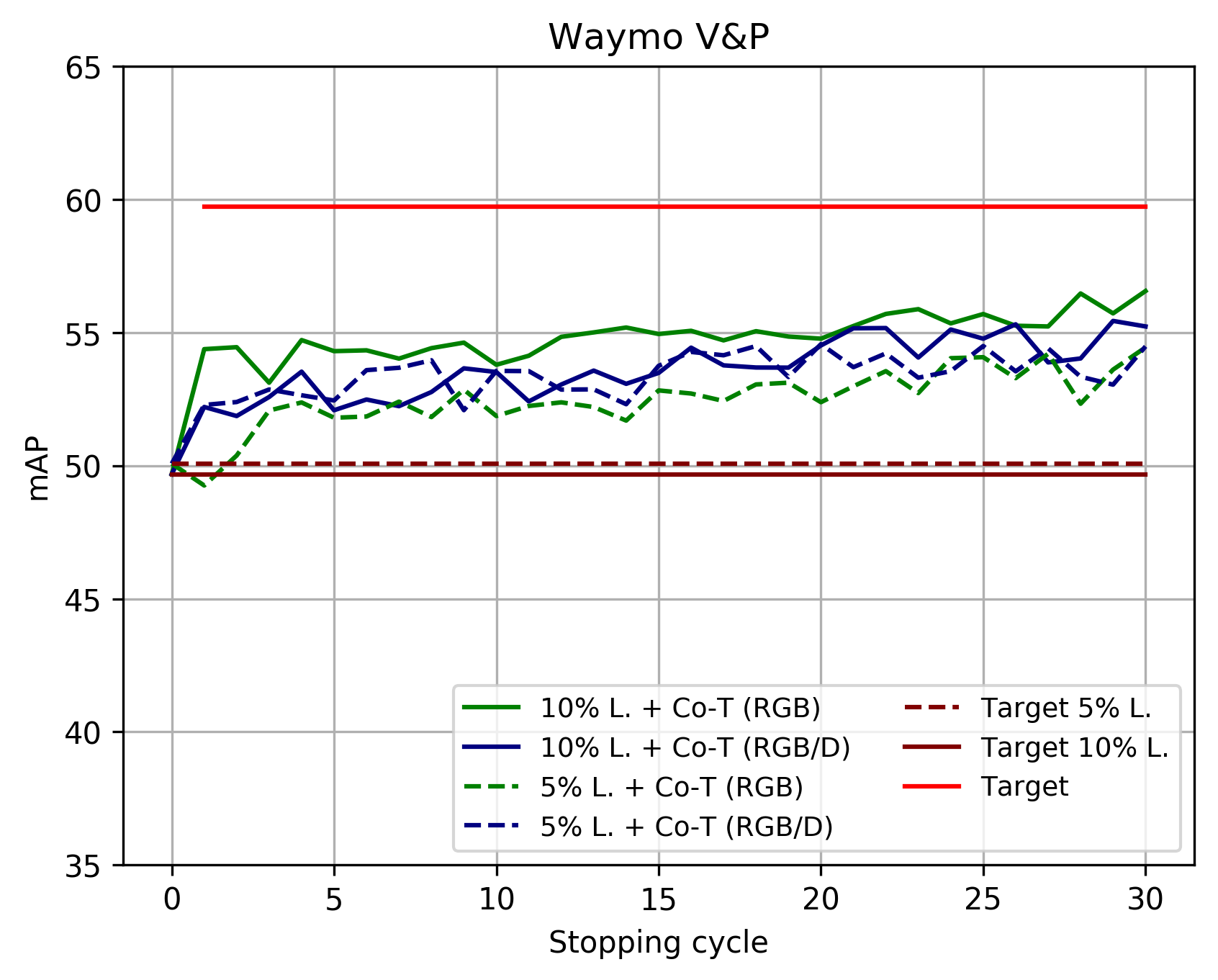}
    \end{subfigure}
\caption{V\&P detection accuracy of co-training approaches as a function of the stopping cycle. Co-T (RGB) and Co-T (RGB/D) refer to single and multi modal co-training, respectively. Target refers to the use of the 100\% labeled training data, while Target {$p$\%} L. indicates a lower percentage $p\in\{5,10\}$ of labeled data available for training. Accordingly, {$p$\%} L. + Co-T (\emph{view}), {\emph{view}~$\in$\{RGB, RGB/D\}}, are combinations of those. These plots complement the results shown in \Tab{sslresults}.}
\label{fig:cycleplots_perc}
\end{figure}
\begin{figure}
\centering
     \begin{subfigure}[b]{0.497\textwidth}
         \centering
            \includegraphics[trim=0 5 0 7, clip, width=\columnwidth]{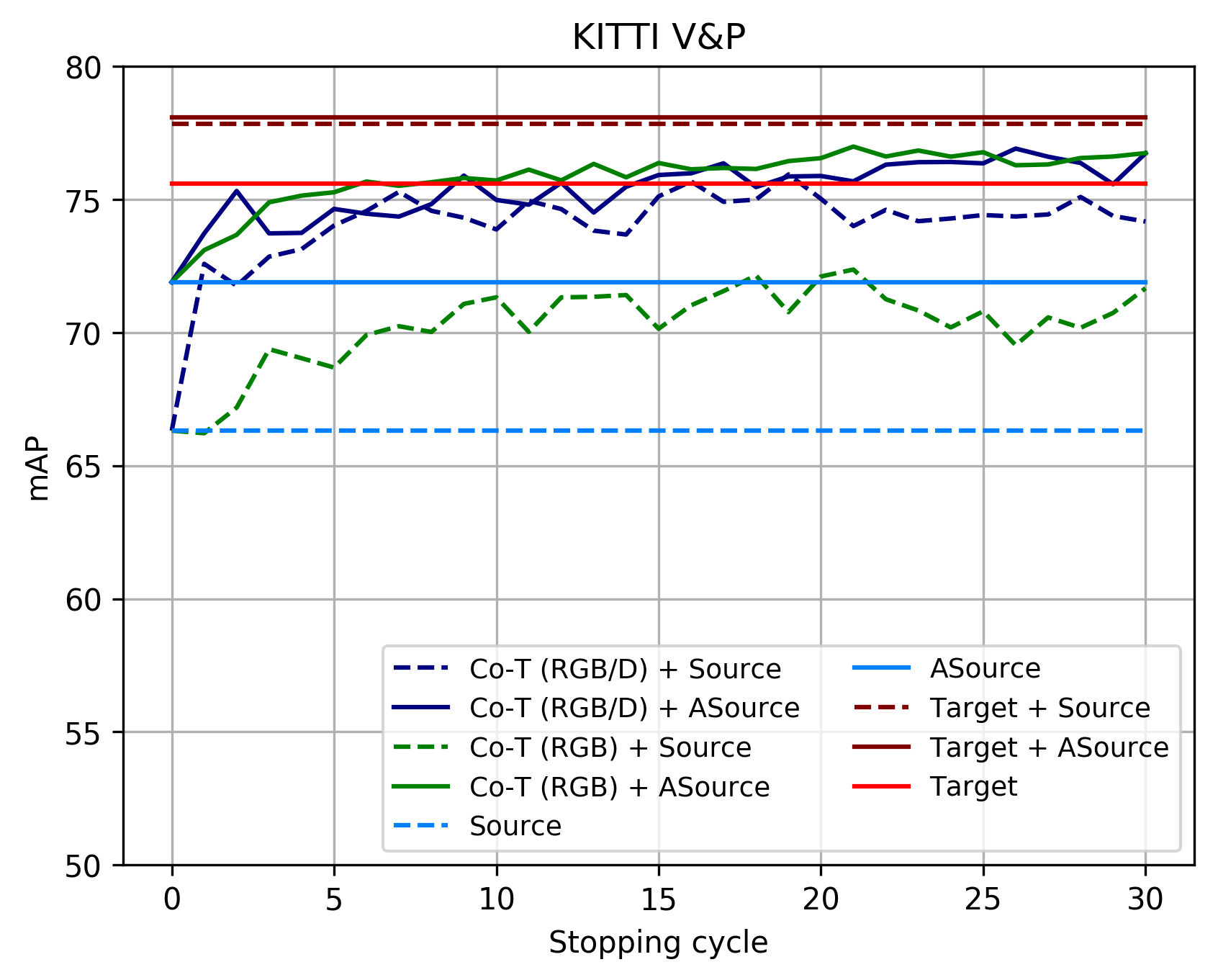}\\
     \end{subfigure}
     \hfill
     \begin{subfigure}[b]{0.497\textwidth}
         \centering
            \includegraphics[trim=0 7 0 5, clip, width=\columnwidth]{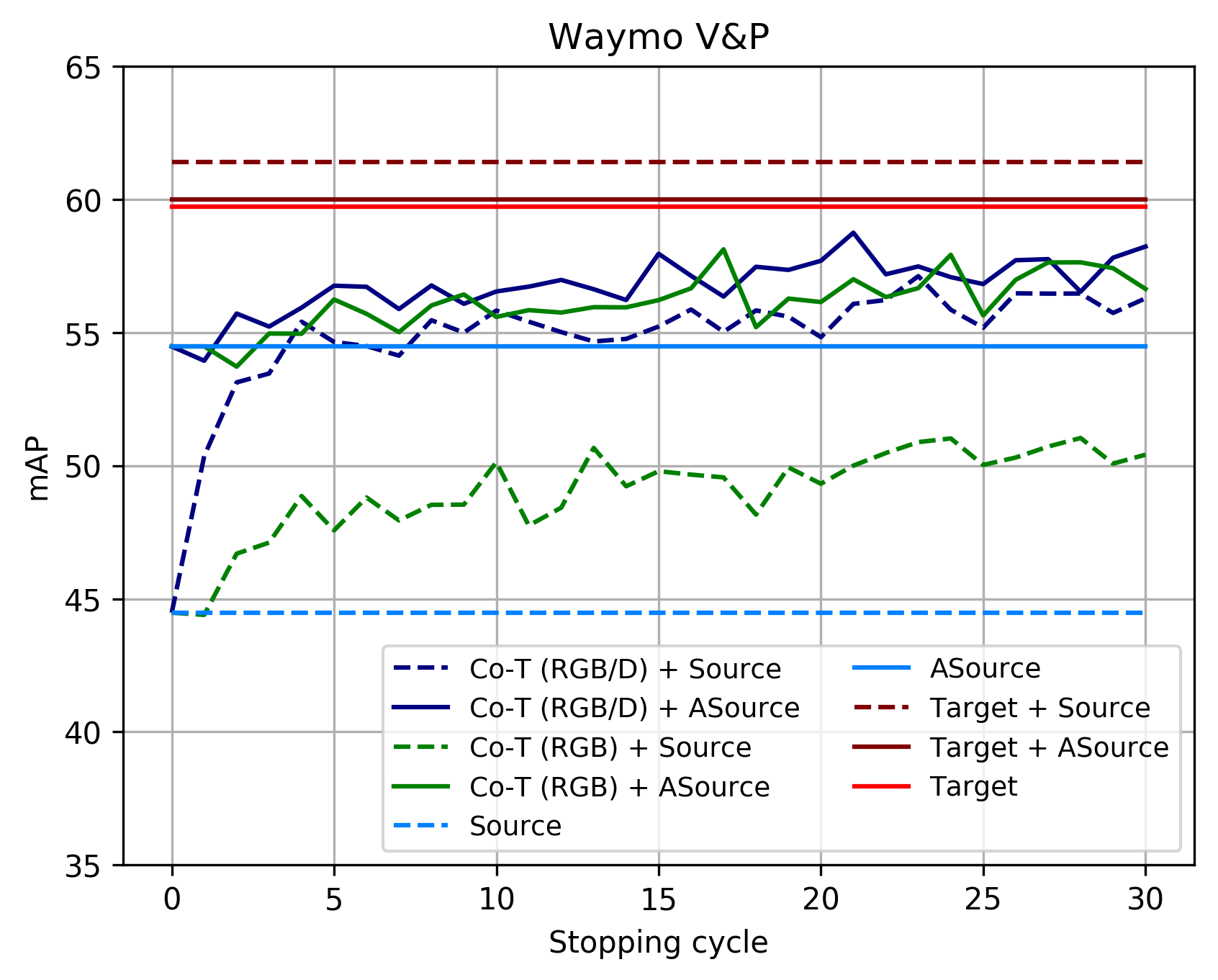}
    \end{subfigure}
\caption{V\&P detection accuracy of co-training approaches as a function of the stopping cycle. These plots are analogous to those in \Fig{cycleplots_perc} for the cases of using virtual-world data, {\ie}, both with domain shift (Source) and reducing it by the use of GANs (ASource). The Targets are the same as in \Fig{cycleplots_perc}. These plots complement the results shown in Tables \ref{tab:Synth} and \ref{tab:Synth2Real}.}
\label{fig:cycleplots_uda}
\end{figure}

\subsubsection{SSL after GAN-based virtual-to-real image translation}
\label{sssec:results-ssl-da}

\Tab{Synth2Real} is analogous to \Tab{Synth}, just changing the original virtual-world images ($\Vds$) by their GAN-based virtual-to-real translated counterparts ($\VGKds/ \VGWds$). In the case of testing on $\Wdstest$ and having $\VGWds$ involved in the training, we apply the BB resizing mentioned in \ssSect{results-ssl-ds} for the quantitative evaluation. Focusing on the V\&P results, we see that both the UB and LB of \Tab{Synth2Real} show higher accuracy than in \Tab{Synth}, which is due to the reduction of the virtual-to-real domain shift achieved thanks to the use of $\VGKds$ / $\VGWds$. Still, co-training enables to improve the accuracy of the LBs, almost reaching the accuracy of the UBs. For instance, in the combined  V\&P detection accuracy, the single-modal co-training is $1.66$ points behind the UB for $\Kds$, and $3.59$ for $\Wds$. Multi-modal co-training is $2.63$ points behind the UB for $\Kds$, and $4.01$ for $\Wds$. Thus, in this case, single-modal co-training is performing better than multi-modal. Therefore, for 2D object detection, we can recommend even single-modal co-training for an SSL scenario where the labeled data comes from a virtual world but a properly trained GAN can perform virtual-to-real domain adaptation. On the other hand, in the case of $\Wds$, co-training from $\VGWds$ gives rise to worse results than by using $\Vds$. We think this is due to a worse depth estimation (see \Fig{monodepth}). In general, this suggests that whenever it is possible, training a specific monocular depth estimator for the unlabeled real-world data may be beneficial for multi-modal co-training (recent advances on vision-based self-supervision for monocular depth estimation \cite{Godard:2019, Gurram:2021} can be a good starting point). For this particular case, training the virtual-to-real domain adaptation GAN simultaneously to the monocular depth estimation CNN could be an interesting idea to explore in the future (we can leverage inspiration from \cite{Zhao:2019GASDA, Pnvr:2020SharinGAN}).

\begin{figure*}
\centering
\includegraphics[width=1\textwidth]{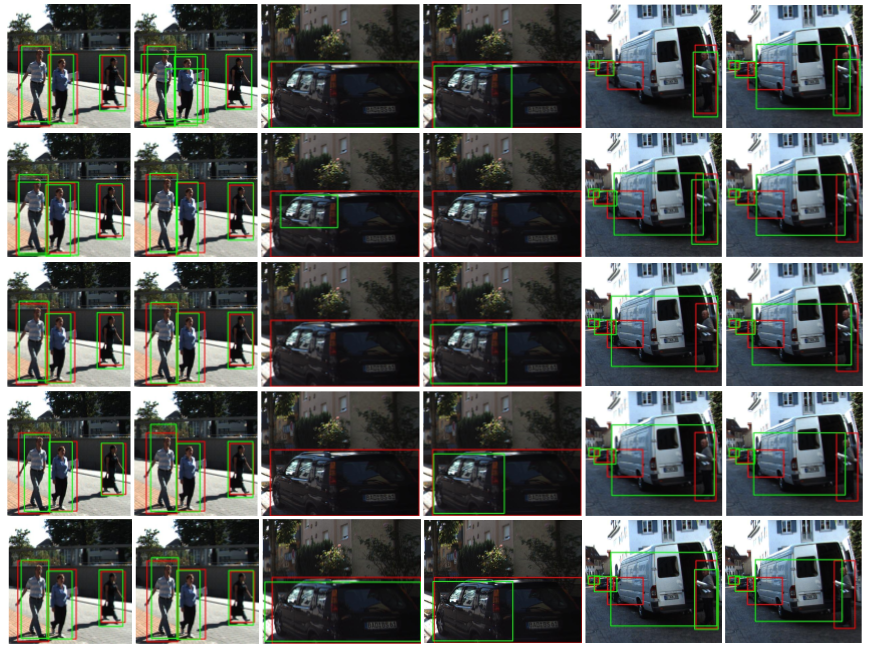}
\caption{Qualitative results of how $\OD_F$ would perform on $\Kdstest$ by stopping co-training at different cycles. We focus on co-training and object detection working from $\VGKds$ (ASource). There are three blocks of results vertically arranged. At each block, the top-left image shows the results when using the 100\% human-labeled training data plus $\VGKds$ (Target + ASource), {\ie}, UB results. Detection results are shown as green BBs, and GT as red BBs. The top-right image of each block shows the results that we would obtain without leveraging the unlabeled data (ASource), {\ie}, LB results. The rest of the rows of the block, from top-second to bottom, correspond to stopping co-training at cycles 1, 10, 20, and automatically. In these rows, the images at the left column correspond to multi-modal co-training ({\ie}, Co-T (RGB/D)) and those at the right column to single-modal co-training ({\ie}, Co-T (RGB)).} 
\label{fig:co-T_epochs_kitti}
\end{figure*}

\begin{figure*}
\centering
\includegraphics[width=1\textwidth]{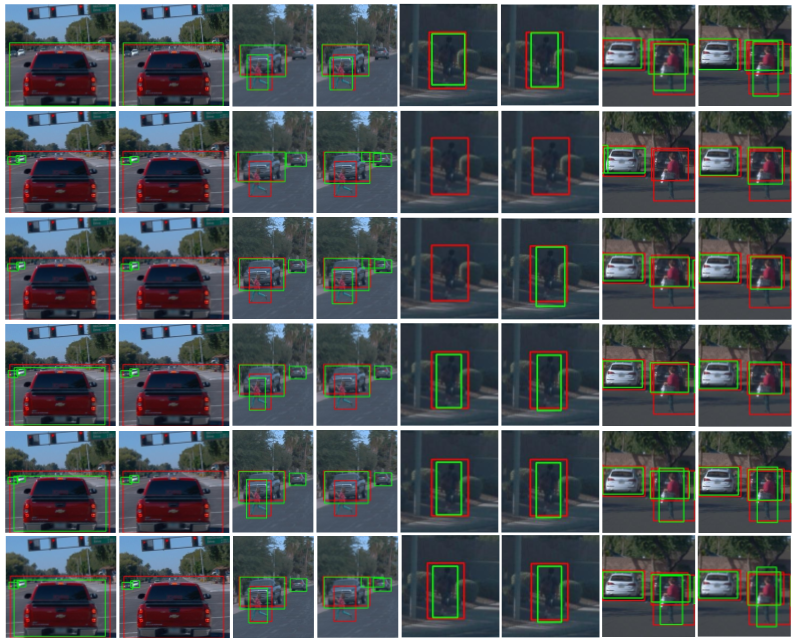}
\caption{Qualitative results similar to those in \Fig{co-T_epochs_kitti}, but testing on $\Wdstest$, co-training from $\Vds$ in the multi-modal case (left column of each block), and $\VGWds$ in the single-modal case (right column of each block). Since, in these examples, the two co-training modalities are based on different (labeled) data, the first row of each block shows the respective UB results, {\ie}, those based on training with $\Wdstrain$ and either with $\Vds$ (left image: Target + Source) or $\VGWds$ (right image: Target + ASource). The second row of each block shows the respective results we would obtain without leveraging the unlabeled data, {\ie}, the LBs based on training with $\Vds$ (left image: Source) or $\VGWds$ (right image: ASource). As in \Fig{co-T_epochs_kitti}, the rest of the rows of each block correspond to stopping co-training at cycles 1, 10, 20, and automatically.}
\label{fig:co-T_epochs_waymo}
\end{figure*}

\subsubsection{Analyzing co-training cycles}
\label{sssec:results-cycles}

Figures \ref{fig:cycleplots_perc} and \ref{fig:cycleplots_uda} illustrate how co-training strategies would perform as a function of the stopping cycle, for a standard SSL setting (\Fig{cycleplots_perc}), as well as under domain shift (Source) and when this is reduced (ASource) by using $\VGKds$ / $\VGWds$ (\Fig{cycleplots_uda}). We take the self-labeled images at different co-training cycles ($x$-axis) as if these cycles were determined to be the stopping ones. The labeled images together with the self-labeled by co-training up to the indicated cycle are used to train the corresponding $\OD_F$. Then, we plot ($y$-axis) the accuracy (mAP) of each $\OD_F$ in the corresponding testing set, {\ie}, either $\Kdstest$ or $\Wdstest$. We can see how co-training strategies allow improving over the LBs from early iterations and, although slightly oscillating, keep improving until stabilization is reached. No drifting to erroneous self-labeling is observed. At this point, the object samples which remain as unlabeled but are required to reach the maximum accuracy, probably are too different in some aspect from the labeled and self-labeled ones ({\eg}, they may be under a too-heavy occlusion) and would never be self-labeled without additional information. Then, combining co-training with active learning (AL) cycles could be an interesting alternative, since occasional human loops could help co-training to progress more. We see also how when the starting point for co-training is at a lower accuracy, multi-modal co-training usually outperforms single-modal  ({\eg}, in the 5\% setting and under domain shift).

\subsubsection{Qualitative results}
\label{sssec:results-qualitative}

Figures \ref{fig:co-T_epochs_kitti} and \ref{fig:co-T_epochs_waymo} present qualitative results for {$\OD_F$'s} trained after stopping co-training at cycles 1, 10, 20 and when it stops automatically ({\ie}, the stopping condition of the loop in \Alg{co-training} becomes true). The shown examples correspond to the most accurate setting for each dataset; {\ie}, for $\Kds$ (\Fig{co-T_epochs_kitti}) this is the co-training from $\VGKds$ no matter the modality, while for $\Wds$ (\Fig{co-T_epochs_waymo}) this is the co-training from $\Vds$ in the multi-modal case and from $\VGWds$ in the single-modal. Note that Tables \ref{tab:sslresults}, \ref{tab:Synth}, and \ref{tab:Synth2Real}, suggest to combine co-training with virtual-world data to obtain more accurate {$\OD_F$'s}. 

In the left block of \Fig{co-T_epochs_kitti}, we show a case where both co-training modalities perform similarly on pedestrian detection, with final detections (green BBs) very close to the GT (red BBs), and clearly better than if we do not leverage the unlabeled data (top-right image of the block). We see also that the results are very similar to the case of using the 100\% of human-labeled data (top-left image of the block). Moreover, even from the initial cycles of both co-training modalities the results are reasonably good, although, the best is expected when co-training finishes automatically (bottom row of the block), {\ie}, after the minimum number of cycles is exceeded ($K_{min}=20$ in \Tab{parameters}). In the mid-block, we see that only multi-modal co-training helps to properly detect a very close and partially occluded vehicle. In the right block, only multi-modal co-training helps to keep and improve the detection of a close pedestrian. Both co-training modalities help to keep an initially detected van, but multi-modal co-training induces a better BB adjustment. This is an interesting case. Since $\Vds$ only contains different types of cars but lacks a meaningful number of van samples, and $\Kds$ only has a very small percentage of those labeled, we have focused our study on the different types of cars. Therefore, vans are neither considered for training nor testing, {\ie}, their detection or misdetection does not affect the mAP metric either positively or negatively. However, co-training is an automatic self-labeling procedure, thus it may capture or keep these samples and then force training with them. Moreover, in this setting, the hard-negatives are mined only from the virtual-world images (translated or not by a GAN) since they are fully labeled. Thus, if no sufficient vans are part of the virtual-world images, these objects cannot act as hard negatives, so that they may be detected or misdetected depending on their resemblance to the targeted objects (here types of cars). We think this is the case here. Thus, this is an interesting consideration for designing future co-training procedures supported by virtual-world data. Alternatively, by complementing co-training with occasional AL cycles, these special false positives could be reported by the human in the AL loop (provided we really want to treat them as false positives). On the other hand, in the same block of results, we see also a misdetection (isolated red BB), which does account for the quantitative evaluation. It corresponds to a rather occluded vehicle which is not detected even when relying on human labeling (top-left image of the block). Finally, note the large range of detection distances achieved for vehicles.  

\begin{figure*}
\centering
\includegraphics[width=1\textwidth]{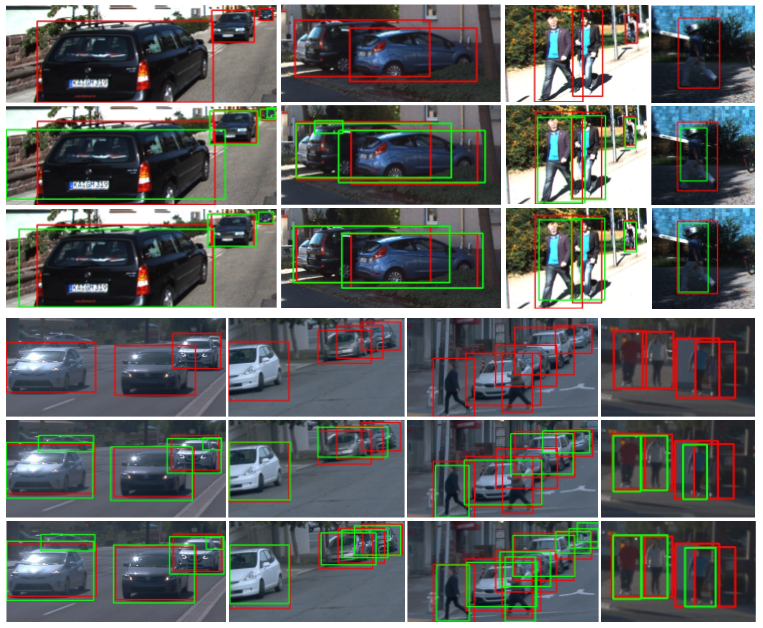}
\caption{Qualitative results on $\Kdstest$ (top block of rows) and $\Wdstest$ (bottom block of rows). In each block, we show (top row) GT as red BBs, (mid row) detections, as green BBs, when training with $\XL$, (bottom row) detections with $\XL\cup\XP$. In this case, $\XP$ comes from applying C-T (RGB/D) on either $\Kdstrain$ or $\Wdstrain$, and $\XL$ is $\VGKds$ for $\Kdstrain$, while it is $\Vds$ for $\Wdstrain$.} 
\label{fig:qualitative_kitti_waymo}
\end{figure*} 

In the left block of \Fig{co-T_epochs_waymo}, we see even a larger detection range for the detected vehicles than in \Fig{co-T_epochs_kitti}. Faraway vehicles (small green BBs) are considered as false positives for the qualitative evaluation because these are not part of the $\Wdstest$ GT (since they do not have labeled 3D BBs from which the 2D BBs are obtained). Thanks to the use of virtual-world data, these vehicles are detected (second row of the block) and both co-training modalities do not damage their detection. Note how the UBs based on virtual-world data and human-labeled real-world data are not able to detect such vehicles (first row of the block) because human labeling did not consider these faraway vehicles, while co-training does consider them as such. Besides, multi-modal co-training enables the detection of the closer vehicle since cycle 10. In the next block to the right, multi-modal co-training enables to detect a close kid since cycle 10, while single-modal does not at the end. In addition, single-modal co-training also introduces a distant false positive. Similarly to the left block, in this block both co-training modalities keep an unlabeled vehicle detected thanks to the use of the virtual-world data (second row), not detected (first row) when these data are complemented with human-labeled data (since, again, this vehicle is not even labeled). What is happening in these cases, is that there is a lack of real-world human-labeled 3D BBs for distant vehicles, which is compensated by the use of virtual-world data and maintained by co-training. In the next block to the right, we see how a pedestrian is detected thanks to both co-training methods since only using virtual-world data was not possible (second row). In the right block, both co-training modalities allow for vehicle and pedestrian detections similar to the UBs (first row). Note that the vehicle partially hidden behind the pedestrian was not detected by only using virtual-world data (second row), and neither was detected the pedestrian when using $\Vds$ (second row, left) or was poorly detected when using $\VGWds$ (second row, right). 
 
Finally, \Fig{qualitative_kitti_waymo} shows additional qualitative results on $\Kdstest$ and $\Wdstest$ when using multi-modal co-training, in the case of $\Kdstest$ based on $\VGKds$ and $\Vds$ for $\Wdstrain$, {\ie}, we show the results of the respective best models. Overall, in the case of $\Kdstest$, we see how multi-modal co-training (Co-T (RGB/D)) enables to better adjust detection BBs, and removing some false positives. In the case of $\Wdstest$, multi-modal co-training enables to keep even small vehicles that are not part of the GT but are initially detected thanks to the use of virtual-world data. It also helps to detect vehicles and pedestrians not detected by only using the virtual-world data, although further improvements are needed since some pedestrians are still difficult to detect even with co-training.

\begin{table}
\centering
\caption{Digging in the results throw three post-processing settings applied to co-training pseudo-labels: (FP) where we remove the false positive pseudo-labels; (BB) where we change the pseudo-labels by the corresponding GT ({\ie}, in terms of Figures \ref{fig:co-T_epochs_kitti}---\ref{fig:qualitative_kitti_waymo}, green BBs are replaced by red ones); (FP+BB) which combines both. This table follows the terminology of Tables \ref{tab:sslresults}, \ref{tab:Synth}, and  \ref{tab:Synth2Real}. $\Delta_{X}$, $X\in\{\mbox{FP,BB,FP+BB}\}$, stands for difference of setting $X$ minus the respective original ({\ie}, using the co-training pseudo-labels).Moreover, for each block of results, we add the \#FP/FP\% row, where \#FP refers to the total number of false positives that are used to train the final object detector, $\OD_F$, while FP\% indicates what percentage they represent regarding the whole set (labeled and self-labeled BBs) used to train $\OD_F$.}\label{tab:Synth2RealNoFalsePositives}

\begin{tabular}{@{}lcccccc@{}}
\toprule
&\multicolumn{3}{c}{$\Xtest=\Kdstest$} & \multicolumn{3}{c}{$\Xtest=\Wdstest$} \\
Training set & V & P & V\&P & V & P & V\&P  \\
\toprule
Target + ASource (UB)                 &  86.82          &  71.59          &  79.21          &  64.56          &  63.44          &  64.00 \\
\midrule
5\% Labeled + Co-T (RGB/D)            &  78.64          &  57.40          &  68.02          &  58.42          &  \textul{56.98} &  57.70 \\
5\% Labeled + Co-T (RGB/D)/FP         &  79.29   &  \textbf{60.50}  &  69.90 &  59.28          &  55.89          &  57.59 \\
5\% Labeled + Co-T (RGB/D)/BB         & \textbf{85.18}  &  \textul{58.87} &  \textbf{72.03} & \textbf{63.25}  &  56.58          &  \textbf{59.92}  \\
5\% Labeled + Co-T (RGB/D)/FP+BB      &  \textbf{85.61} &  58.75 &  \textbf{72.18} &  \textul{62.49} &  \textbf{57.63} &  \textbf{60.06} \\
$\Delta_{\mbox{FP}}$                  &  +0.65          &  +3.10          &  +1.88          &  +0.86          &  -1.09          &  -0.11 \\
$\Delta_{\mbox{BB}}$                  &  +6.54          &  +1.47          &  +4.01          &  +4.83          & -0.40           &  +2.22  \\
$\Delta_{\mbox{FP+BB}}$               &  +6.97          &  +1.35          &  +4.16          &  +4.07          &  +0.91          &  +2.36 \\
 \#FP/FP\%                              &  1723/13.65\%   &  275/16.33\%    &                 &  5952/10.39\%   & 731/11.33\%    &   \\

\midrule
10\% Labeled + Co-T (RGB/D)           &  79.68          &  60.55 &  70.12          &  59.54          &  \textbf{57.17} &  58.36 \\
10\% Labeled + Co-T (RGB/D)/FP        &  79.81  &  \textbf{61.65} &  70.73 &  60.24 &  \textbf{57.38} &  58.81 \\
10\% Labeled + Co-T (RGB/D)/BB        & \textbf{85.28}   &  59.21  &  \textbf{72.25}  &  \textbf{63.01}  &    56.07        &  \textul{59.54}  \\
10\% Labeled + Co-T (RGB/D)/FP+BB     &  \textul{83.23} &  \textul{61.03}          &  \textbf{72.13} &  \textbf{63.20} &  \textbf{56.99} &  \textbf{60.10} \\
$\Delta_{\mbox{FP}}$                  &  +0.13          &  +1.10          &  +0.61          &  +0.70          &  +0.21          &  +0.45 \\
$\Delta_{\mbox{BB}}$                  &    +5.60        &      -1.34       &    +2.01         &     +3.47        &  -1.1           &   +1.18 \\
$\Delta_{\mbox{FP+BB}}$               &  +3.55          &  +0.48          &  +2.01          &  +3.66          &  -0.18          &  +1.74 \\
 \#FP/FP\%                                      &  1998/14.06\%   &  408/16.83\%          &            &  4553/7.42\%          & 547/7.30\%           &   \\
\midrule
Co-T (RGB/D) + Source                 &  82.90          &  67.36          &  75.13          &  \textbf{64.40} &  \textbf{59.17} &  \textbf{61.79} \\
Co-T (RGB/D) + Source/FP              &  83.37 &  \textul{70.95} &  \textul{77.16} &  57.68          &  56.04          &  56.86 \\
Co-T (RGB/D) + Source/BB              &  \textbf{88.94}  &  61.69  &   75.32 &  62.14  &   \textul{57.22}         &  \textul{59.68}  \\
Co-T (RGB/D) + Source/FP+BB           &  \textbf{89.07} &  \textbf{71.88} &  \textbf{80.48} & \textul{62.56}  &  56.59 &  59.58 \\
$\Delta_{\mbox{FP}}$                  &  +0.47          &  +3.59          &  +2.03          &  -6.72          &  -3.13          &  -4.93 \\
$\Delta_{\mbox{BB}}$                  &     +6.04        &       -5.67      &      +0.19       &     -2.26        &  -1.95           &   -2.11 \\
$\Delta_{\mbox{FP+BB}}$               & +6.17           & +4.52           &  +5.35          & -1.84           &  -2.58          &  -2.21\\
 \#FP/FP\%                &  3281/3.57\%          &  883/0.87\%          &           &  18293/21.06\%          & 970/2.02\%           &   \\
\midrule
Co-T (RGB/D) + ASource                &  83.68 &  69.48          &  76.58 &  61.49 &  \textbf{59.33} &  \textbf{60.41} \\
Co-T (RGB/D) + ASource/FP             &  83.07          &  \textul{70.40}  &  76.74 &  60.67          &  \textul{57.72} &  59.20 \\
Co-T (RGB/D) + ASource/BB             & \textbf{89.27}   & 68.63   &  \textul{78.95}  &  \textul{62.06}  &   57.64         & \textul{59.85}   \\
Co-T (RGB/D) + ASource/FP+BB          &  \textbf{88.93} &  \textbf{71.45} &  \textbf{80.19} &  \textbf{64.67} &  55.27          &  \textbf{59.97} \\
$\Delta_{\mbox{FP}}$                  &  -0.61          &  +0.92          &  0.16           &  -0.82          &  -1.61          &  -1.21 \\
$\Delta_{\mbox{BB}}$                  &    +5.59         &     -0.85        &   +3.61          &     +0.57        &  -1.69           &   -0.56 \\
$\Delta_{\mbox{FP+BB}}$               &  +5.25          &  +1.97          &  +3.61          &  +3.18          &  -4.06          &  -0.44 \\
\#FP/FP\%                &  3097/5.59\%          &  479/1.03\%          &            &  20949/23.15\%          & 816/1.70\%           &   \\               
\bottomrule
\end{tabular}
\end{table}

\subsubsection{Answering (Q1) and (Q2)}
\label{sssec:results-Q1-Q2}

After presenting our multi-modal co-training and the extensive set of experiments carried out, we can answer the research questions driving this study. In particular, we base our answers in the quantitative results presented in Tables \ref{tab:sslresults}, \ref{tab:Synth}, \ref{tab:Synth2Real}, the plots shown in Figures \ref{fig:cycleplots_perc} and \ref{fig:cycleplots_uda}, as well as the qualitative examples shown in Figures \ref{fig:co-T_epochs_kitti}, \ref{fig:co-T_epochs_waymo}, and \ref{fig:qualitative_kitti_waymo}, together with the associated comments we have drawn from them.

(Q1) \emph{Is multi-modal (RGB/D) co-training effective on the task of providing pseudo-labeled object BBs?} Indeed, multi-modal co-training is effective for self-labeling object BBs under different settings, namely, for standard SSL (no domain shift, a few human-labeled data) and when using virtual-world data (many virtual-world labeled data, but no human-labeled data) both under domain shift and after reducing it by GAN-based virtual-to-real image translation. The achieved improvement over the lower bound configurations is significant, allowing to be almost in pair with upper bound configurations. In the standard SSL setting, by only labeling the 5\% of the training dataset, multi-modal co-training allows obtaining accuracy values relatively close to the upper bounds. When using virtual-world data, {\ie}, without human labeling at all, the same observations hold. Moreover, multi-modal co-training and GAN-based virtual-to-real image translation have been shown to complement each other.

(Q2) \emph{How does perform multi-modal (RGB/D) co-training compared to single-modal (RGB)?} We conclude that in a standard SSL setting (no domain shift, a few human-labeled data) and under virtual-to-real domain shift (many virtual-world labeled data, no human-labeled data) multi-modal co-training outperforms single-modal. In the latter case, when GAN-based virtual-to-real image translation is performed both co-training modalities are on pair; at least, by using an off-the-shelf monocular depth estimation model not specifically trained on the translated images. 

To drive future research, we have performed additional experiments. These consist in correcting the pseudo-labels obtained by multi-modal co-training in three different ways, namely, removing false positives (FP), adjusting the BBs to the ones of the GT (BB) for correctly self-labeled objects (true positives), and a combination of both (FP+BB). After changing the pseudo-labels in that way, we train the corresponding $\OD_F$ models and evaluate them. \Tab{Synth2RealNoFalsePositives} presents the quantitative results. Focusing on the standard SSL setting (5\%, 10\%), we see that the main problem for vehicles in $\Kds$ is BB adjustment, while for pedestrians is the introduction of FPs. In the latter case, false negatives (FN; {\ie}, missing self-labeled objects) seem to be also an issue to reach upper bound accuracy. When we have the support of virtual-world data, FNs do not seem to be a problem, and addressing BB correction for vehicles and removing FPs for pedestrians would allow reaching upper bounds. In the case of $\Wds$, we came to the same conclusions for vehicles, the main problem is BB adjustment, while in the case of pedestrians the main problem is not that clear. In other words, there is more balance between FP and BB. On the other hand, regarding these additional experiments, we trust more the conclusions derived from $\Kds$. The reason is that, as we have seen in Figures \ref{fig:co-T_epochs_waymo} and \ref{fig:qualitative_kitti_waymo}, co-training was correctly self-labeling objects that are not part of the GT, so in this study, these are either considered FPs and so wrongly removed (FP, FP+BB settings), or would not have a GT BB to which adjust them  (BB, FP+BB settings).

After this analysis, we think we can explore two main future lines of research. First, to improve BB adjustment, we could complement multi-modal co-training with instance segmentation, where using Mask R-CNN \cite{He:2017} would be a natural choice. Note that virtual-world data can also have instance segmentation as part of their GT suite. Second, to remove FPs, we could add an AL loop where humans could remove even several FP with a few clicks (note that this is much easier than delineating object BBs). On the other hand, additional CNN models could be explored to avoid FPs as a post-processing step to multi-modal co-training. Besides these ideas, we think that, whenever is possible, the monocular depth estimation model should be trained on the target domain data, rather than trying to use an off-the-shelf model. Since we think that not doing so was damaging the combination of multi-modal co-training and GAN-based virtual-to-real image translation, an interesting approach would be to perform both tasks simultaneously.

\section{Conclusions}
\label{sec:c}

In this paper, we have addressed the curse of data labeling for onboard deep object detection. In particular, following the SSL paradigm, we have proposed multi-modal co-training for object detection. This co-training relies on a data view based on appearance (RGB) and another based on estimated depth (D), the latter obtained by applying monocular depth estimation, so keeping co-training as a single-sensor method. We have performed an exhaustive set of experiments covering the standard SSL setting (no domain shift, a few human-labeled data) as well as the settings based on virtual-world data (many virtual-world labeled data, no human-labeled data) both with domain shift and without (using GAN-based virtual-to-real image translation). In these settings, we have compared multi-modal co-training and appearance-based single-modal co-training. We have shown that multi-modal co-training is effective in all settings. 
In the standard SSL setting, from a 5\% of human-labeled training data, co-training can already lead to a final object detection accuracy relatively close to upper bounds ({\ie}, with the 100\% of human labeling). The same observation holds when using virtual-world data, {\ie}, without human labeling at all. Multi-modal co-training outperforms single-modal in standard SSL and under domain shift, while both co-training modalities are on pair when GAN-based virtual-to-real image translation is performed; at least, by using an off-the-shelf depth estimation model not specifically trained on the translated images. Moreover, multi-modal co-training and GAN-based virtual-to-real image translation have been proved to be complementary.
%
%
For the future, we plan several lines of work, namely, improving the adjustment of object BBs by using instance segmentation upon detection and removing false-positive pseudo-labels by using a post-processing AL cycle. Moreover, we believe that the monocular depth estimation model should be trained based on target domain data whenever possible. When GAN-based image translation is required, we could jointly train the monocular depth estimation model and the GAN on the target domain. Besides, we would like to extend co-training experiments to other classes of interest for onboard perception (traffic signs, motorbikes, bikes, etc.), as well as adapting the method to tackle other tasks such as pixel-wise semantic segmentation.

\bibliographystyle{ieee}
\bibliography{references}

\end{document}